\documentclass[lettersize,journal]{IEEEtran}
\usepackage{amsmath,amsfonts}
\usepackage{algorithmic}
\usepackage{algorithm}
\usepackage{array}
\usepackage[caption=false,font=normalsize,labelfont=sf,textfont=sf]{subfig}
\usepackage{textcomp}
\usepackage{stfloats}
\usepackage{url}
\usepackage{verbatim}
\usepackage{graphicx}
\usepackage{booktabs}
\usepackage{cite}
\usepackage{caption}
\usepackage{float}
\usepackage{amsmath}
\usepackage{multirow}
\usepackage{colortbl}
\usepackage{amssymb} 

\usepackage{amsmath}

\DeclareMathOperator*{\argmin}{arg\,min}

\usepackage{graphicx}


\begin{document}

\title{Training and Tuning Generative Neural Radiance Fields for Attribute-Conditional 3D-Aware Face Generation}



\author{Jichao~Zhang,
        Aliaksandr~Siarohin,
        Yahui~Liu,
        Hao~Tang, 
        Nicu~Sebe~\IEEEmembership{Senior~Member,~IEEE} and 
        Wei~Wang
\IEEEcompsocitemizethanks{\IEEEcompsocthanksitem Jichao Zhang is with the School of Computer Science, Ocean University of China, Shandong, China. E-mail: zhang163220@gmail.com. 
\IEEEcompsocthanksitem  Aliaksandr Siarohin is with the Snap Research, Santa Monica, CA,
US. E-mail: aliaksandr.siarohin@gmail.com.
\IEEEcompsocthanksitem  Yahui Liu is a Principal Engineer in Huawei, Shenzhen, China. E-mail: yahui.cvrs@gmail.com.
\IEEEcompsocthanksitem Hao Tang is with the National Key Laboratory for Multimedia Information Processing, 
School of Computer Science, Peking University, China. E-mail: bjdxtanghao@gmail.com.
\IEEEcompsocthanksitem Wei Wang is with the Institute of Information Science, Beijing Jiaotong University, Beijing, China. E-mail: wangwei1990@gmail.com.
\IEEEcompsocthanksitem Nicu Sebe is with the Department of Information Engineering and Computer Science (DISI), University of Trento, Italy. E-mail: sebe@disi.unitn.it. 
}

}

\markboth{}%
{Shell \MakeLowercase{\textit{et al.}}: A Sample Article Using IEEEtran.cls for IEEE Journals}


\maketitle

\begin{abstract}
Generative Neural Radiance Fields (GNeRF)-based 3D-aware GANs have showcased remarkable prowess in crafting high-fidelity images while upholding robust 3D consistency, particularly face generation. However, specific existing models prioritize view consistency over disentanglement, leading to constrained semantic or attribute control during the generation process. While many methods have explored incorporating semantic masks or leveraging 3D Morphable Models (3DMM) priors to imbue models with semantic control, these methods often demand training from scratch, entailing significant computational overhead. In this paper, we propose a novel approach: a conditional GNeRF model that integrates specific attribute labels as input, thus amplifying the controllability and disentanglement capabilities of 3D-aware generative models. Our approach builds upon a pre-trained 3D-aware face model, and we introduce a Training as Init and Optimizing for Tuning (TRIOT) method to train a conditional normalized flow module to enable the facial attribute editing, then optimize the latent vector to improve attribute-editing precision further. Our extensive experiments substantiate the efficacy of our model, showcasing its ability to generate high-quality edits with enhanced view consistency while safeguarding non-target regions. The code for our model is publicly available at \url{https://github.com/zhangqianhui/TT-GNeRF}.
\end{abstract}


\section{Introduction}
High-quality image generation and semantic disentanglement are long-standing goals in computer vision and computer graphics. In recent years, Generative Adversarial Networks (GANs)~\cite{gf2014} and their variants have received significant attention for their ability to produce high-quality image generation and editing. These methods have greatly improved visual fidelity, rendering speed, and interactive controls compared to traditional computer graphics pipelines.

Numerous previous works~\cite{choi2018stargan,abdal2021styleflow,tewari2020stylerig,shi2021lifting} have focused on realistic face editing. These approaches rely on image-to-image translation models~\cite{choi2018stargan,liu2019stgan} or leverage the disentanglement abilities~\cite{abdal2021styleflow,tewari2020stylerig,shi2021lifting} of StyleGAN~\cite{karras2019style,karras2020analyzing}. These methods can be broadly classified into supervised and unsupervised categories. Unsupervised methods typically search for interpretable directions using PCA~\cite{harkonen2020ganspace} or introduce soft orthogonality constraints~\cite{he2021eigengan,voynov2020unsupervised} in the latent space. However, these approaches provide only coarse controls. Supervised methods~\cite{choi2018stargan,abdal2021styleflow}, on the other hand, utilize specific attribute labels as conditions. However, these methods lack precise control over 3D factors such as camera pose because they overlook the underlying 3D scene rendering process. To address this issue, some works~\cite{tewari2020stylerig,deng2020disentangled} have integrated 3D Morphable Face Models (3DMM)~\cite{paysan20093d} to enable control over 3D face pose and facial expression. However, these approaches still suffer from significant challenges, such as view inconsistency and unrealistic texture distortion when poses are drastically varied.

\begin{figure*}[!ht]\small
\centering
\includegraphics[width=1.0\linewidth]{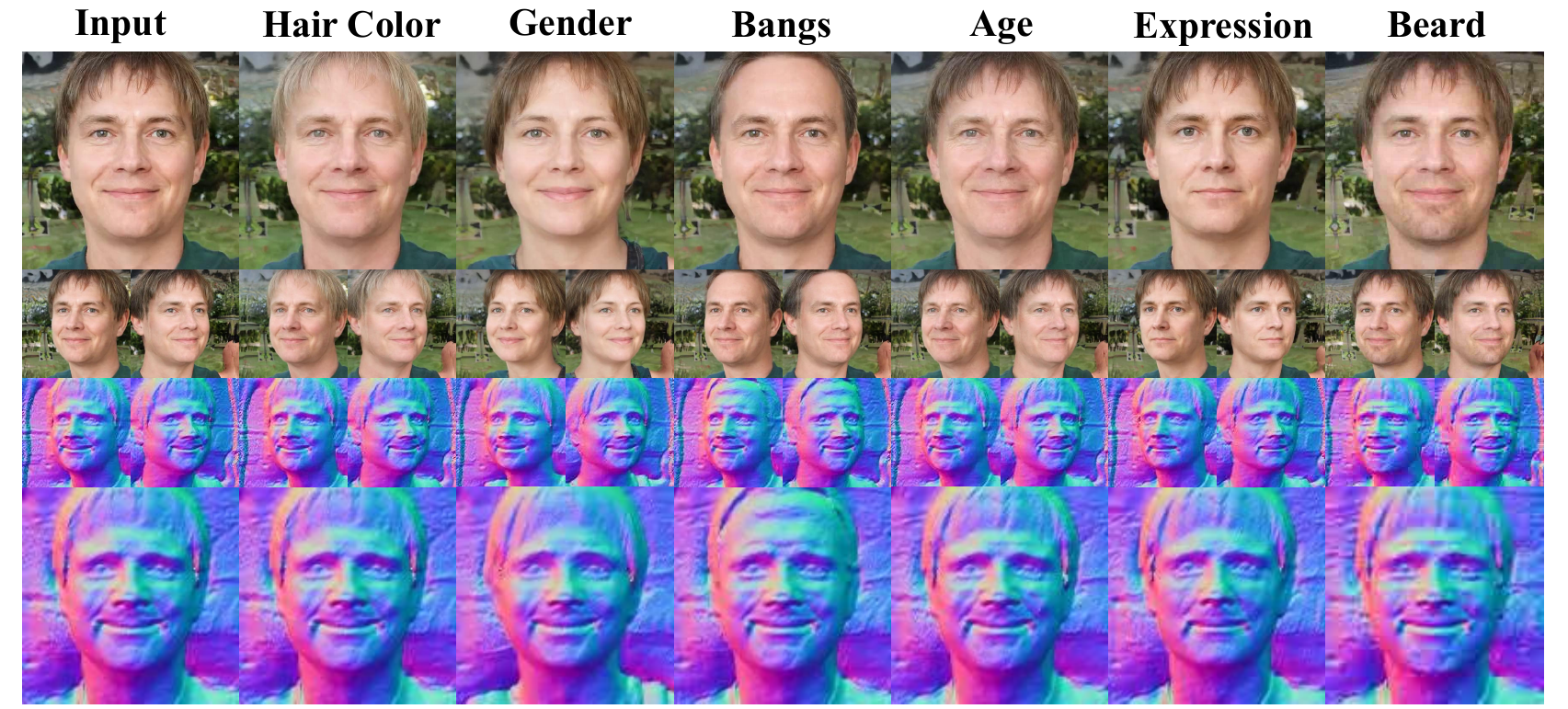}
\caption{Our method produces controllable 3D-aware face generation (first two rows) given specific attributes as guidance and the corresponding normals (bottom two rows). As shown in the normal images, the geometry has been preserved for the attribute ``Hair Color'', while the mouth region of the ``Smiling'' mesh has changed.}
\vspace{-0.4cm}
\label{fig:exp1}
\end{figure*}

Recently, neural radiance fields (NeRF)~\cite{mildenhall2020nerf} have attracted much attention due to their impressive results in novel view-rendering tasks. Specifically, NeRF represents a scene using a continuous function parameterized by a multi-layer perceptron (MLP) that maps a 3D position and a viewing direction to density and radiance values. Since then, many works have been proposed to improve NeRF~\cite{zhang2020nerf++,mueller2022instant} and apply it to various downstream tasks, such as human body modeling~\cite{peng2021neural} and large scene modeling~\cite{tancik2022block}. 

Some 3D-aware image generation methods~\cite{yen2020inerf,chanmonteiro2020pi-GAN} combine NeRF with generative models by extending neural radiance fields with latent conditioning, called Generative Neural Radiance Fields (GNeRF). 3D coordinates are sampled from random camera poses and used as input to an implicit function with latent codes. This function predicts the density and the RGB color. However, these methods are compute-intensive and memory inefficient because they sample many rays in the entire 3D-volume space and require a feed-forward process for each point. They are limited to low-resolution and low-quality generation. To improve generation quality and view-consistency, many approaches~\cite{orel2021stylesdf,chan2021efficient,zhang2022multi} borrow ideas from StyleGAN and integrate the `Style-modules' into the implicit function (\emph{e.g.}, SIREN~\cite{chanmonteiro2020pi-GAN}) or neural rendering module. Some novel algorithms and losses have been carefully designed for 3D-aware generation, such as tri-planes~\cite{chan2021efficient} or multiple-view warping loss~\cite{zhang2022multi}. Although these models create high-quality, view-consistent images, they lack control and disentangling abilities. As explained in VolumeGAN~\cite{xu20213d}, some models are limited to local receptive fields with MLPs, and extracting global structures from their internal representation is difficult. Thus, VolumeGAN utilizes a 3D feature volume module for querying coordinate descriptors, enabling independent controls on the texture and structure factors. However, VolumeGAN still faces quality and view consistency challenges and does not support attribute controls for face manipulation. Recently, some papers~\cite{sun2022fenerf,sun2022controllable,sun2022next3d,sun2022ide,gao2023sketchfacenerf} utilize the 3DMM mesh or semantic masks as a condition to edit or control the 3D face avatar. Mesh-guided methods~\cite{sun2022next3d,sun2022controllable} prioritize expression and head pose transfer but may struggle with appearance editing like ``Hair Color'' manipulation. Mask-guided methods~\cite{sun2022ide,sun2022fenerf,gao2023sketchfacenerf} excel in local attribute editing. However, they may be less effective for global attribute manipulation. An important consideration for both mesh-guided and mask-guided methods is the requirement to train models from scratch, which demands significant computational resources. This high computational cost is an important aspect to keep in mind when considering the feasibility and scalability of these methods.

\begin{figure*}[!ht]\small
\centering
\includegraphics[width=1.0\linewidth]{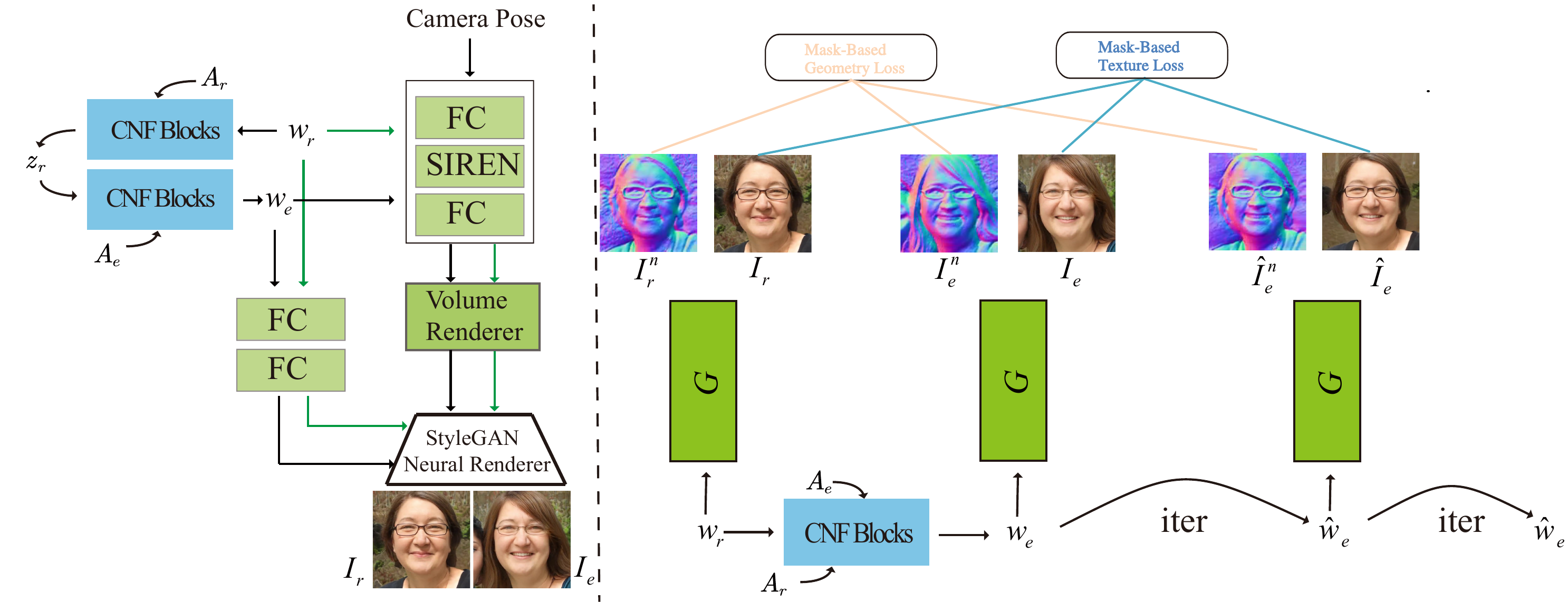}
\caption{``Training as Init, Optimizing for Tuning'' method overview (Attribute ``Expression'' as an example). First, given the pretrained StyleSDF~\cite{orel2021stylesdf},  we follow StyleFlow to train the continuous normalized flows (CNF) to learn the conditional distribution of latent code $w$ (Left). The attribute editing can be enabled by given target labels $A_{e}$ to manipulate the latent code. Second, we regard the edited latent code $w_{e}$ as the initial result, then iteratively optimize the latent code $w_{e}$ to search for a better one by using proposed mask-based geometry and texture loss (Right).}
\vspace{-0.4cm}
\label{fig:model}
\end{figure*}

In this paper, we introduce an attribute-conditional 3D-aware generative model aimed at controlling facial attributes while addressing existing challenges. Our approach offers two key advancements compared to prior methods. Firstly, we leverage attribute labels as conditions for attribute editing tasks, supporting global and local attribute editing. Secondly, we employ a pre-trained 3D-aware GNeRF model as the backbone, eliminating the need for extensive retraining, thus reducing computational resources and time. We propose the ``Training-As-Init, Optimizing-for-Tuning'' (TRIOT) method (Fig.~\ref{fig:model}) to achieve latent disentanglement for precise facial attribute editing. TRIOT consists of two main steps. Initially, we adopt a methodology similar to previous 2D techniques~\cite{abdal2021styleflow} to train a conditional normalizing flow, enabling the learning of latent space distributions and facilitating multiple attribute editing via label input. Subsequently, we utilize the edited latent vector corresponding to target attributes as initialization. This initialized latent vector is then optimized using our proposed semantic-guided texture and geometry consistency losses to enhance editing precision and preserve non-target regions in images.

Furthermore, to explore geometry and texture disentanglement, we introduce an unsupervised optimization method to enable reference-based facial geometry transfer tasks (Fig.~\ref{fig:model2}). This method enhances the ability to transfer facial geometry while maintaining texture consistency, thereby broadening the scope of applications for our model. By combining these advances, our model offers a robust framework for precise attribute editing in facial images, with efficient resource utilization and enhanced disentanglement capabilities.

In summary, the main contributions of this work are:
\begin{itemize}
  \item[1)] We propose a ``Training-as-Init, Optimizing-for-Tuning'' (TRIOT) method, integrating the module-training and latent-optimization method for disentangling latent space and enabling attribute-editing tasks. 
  \item[2)] Our method is flexible and general, and it can be seamlessly integrated into most 3D-Aware GAN architectures.
  \item[3)] Compared to the baseline models, our model achieves high-quality editing with improved view consistency while preserving non-target regions.
 \item[4)] To explore geometry and texture disentanglement, we propose an unsupervised optimization method for the reference-based geometry transfer task.
\end{itemize}

\section{Related work}

\noindent \textbf{Generative Neural Radiance Fields for 3D-Aware Face Generation.} 
Neural radiance fields (NeRF)~\cite{mildenhall2020nerf}, a continuous neural mapping from a 3D position and a 2D viewing direction to the RGB value and density that allows 3D scene modeling and high-quality novel view synthesis. Recently, several NeRF-based methods were proposed to improve rendering speed~\cite{Reiser2021ICCV,barron2021mip,mueller2022instant} and rendering quality~\cite{Niemeyer2021Regnerf,barron2022mipnerf360,verbin2021refnerf}. Moreover, NeRF also promotes the development of many computer-graphics applications, such as human body modeling~\cite{peng2021neural,peng2021animatable}, 3D-aware face generation and editing~\cite{GIRAFFE,orel2021stylesdf,gu2021stylenerf,xu20213d,chan2021efficient,ma2022neural,abdal20233davatargan,sun2022fenerf,zhou2023mate3d,xie2023navinerf,hyung2023local}, large scene modeling~\cite{tancik2022block}, and pose estimation~\cite{yen2020inerf}.

Generative neural radiance fields (GNeRF) are a conditional variant of NeRF, which combines NeRF with GANs to condition the rendering process on a latent code that governs the object's appearance and
shape~\cite{yen2020inerf,chanmonteiro2020pi-GAN,GIRAFFE}. For example, GRAF~\cite{Schwarz2020NEURIPS} achieves this goal by incorporating shape and appearance codes as input. GRAF~\cite{Schwarz2020NEURIPS} achieves better visual fidelity and view consistency than the previous voxel- and feature-based methods~\cite{henzler2019escaping,nguyen2019hologan}. Michael \emph{et al.}~\cite{GIRAFFE} propose the compositional neural feature fields (GIRAFFE) that extend GRAF into 3D-aware multiple-object scene representations. Although GRAF and GIRAFFE can control texture and camera pose, they are limited to low-resolution results and fail to preserve multi-view consistency. Many works~\cite{chanmonteiro2020pi-GAN,zhou2021CIPS3D,deng2021gram,pan2021shading,gu2021stylenerf,orel2021stylesdf,chan2021efficient,zhang2022multi,xue2022giraffe,Schwarz2022,epigraf,xiang2022gram} are trying to address these problems, and most of them inherit the ``image-as-style'' idea from StyleGAN~\cite{karras2020analyzing}. Yang \emph{et al.}~\cite{xue2022giraffe} extend the GIRAFFE to work with high-resolution data. However, this model still suffers from the view-inconsistency problems. Pi-GAN~\cite{chanmonteiro2020pi-GAN} proposes a SIREN module with periodic activation functions. It conditions the style code by feature-wise linear modulation (FILM). The SIREN modules significantly boost image quality and view consistency. To reduce the high computational costs of the volume rendering in Pi-GAN, some models, such as StyleSDF~\cite{orel2021stylesdf}, MVCGAN~\cite{zhang2022multi} and EG3D~\cite{chan2021efficient} propose a hybrid rendering approach. Specifically, they learn a coarse feature field, render it into a low-resolution feature map, and then utilize a style-based 2D network as a ``super-resolution'' module to refine the features for a final high-resolution image. In order to improve view consistency, StyleSDF models signed distance fields, while MVCGAN uses explicit multi-view consistency loss. On the other hand, ED3D proposes a hybrid 3D tri-plane representation. Unlike the mentioned works, CIPS-3D~\cite{zhou2021CIPS3D} keeps the resolution of intermediate feature fields the same as the resolution of the final images. These models can achieve incredible quality generation with strong view consistency, but they cannot edit structures and textures. 

Some research has focused on the disentangling abilities of 3D-aware models. VolumeGAN~\cite{xu20213d} tries to separate shape from texture, while ShadeGAN~\cite{pan2021shading} and NeRFFaceLighting~\cite{jiang2023nerffacelighting} disentangle the light from the albedo. However, they only focus on global factors, such as illumination and textures, and cannot handle more specific attributes, such as hair color and gender. 

Recent methods also explore the semantic control of 3D-aware face model and can be categorized into two types: mesh-guided, and mask-guided (or 2D sketches). Mesh-guided Methods~\cite{sun2022next3d,sun2022controllable} prioritize expression and head pose transfer but may struggle with appearance editing like ``Hair Color" manipulation. On the other hand, mask (or 2D sketches)-guided methods~\cite{sun2022ide,sun2022fenerf,gao2023sketchfacenerf} excel at local attribute editing but may not be as effective for global attributes. It's worth noting that both mesh-guided and mask-guided methods often necessitate training models from scratch. Our proposed method addresses these limitations by disentangling the latent space of a pre-trained 3D-aware GAN, enabling both global and local attribute editing. In a similar vein, LatentSwap3D~\cite{simsar2023latentswap3d} also facilitates attribute editing within the latent space of pre-trained 3D-aware GANs. This method employs a random forest classifier to rank features and identify dimensions controlling specific attributes. By swapping the most relevant features, it accomplishes attribute editing tasks. However, it suffers from low editing intensity, particularly in attributes like "Hair Color" and "Gender". The discussion and comparison with LatentSwap3D can be found at Section~\ref{sec4.2}.


\noindent \textbf{Image-to-Image Translation Architectures for Face Editing.} 
Image-to-Image translation models, \emph{i.e.}, Pix2Pix~\cite{pix2pix2017} and CycleGAN~\cite{CycleGAN2017}, utilize the autoencoder as a generator that has been widely adopted for a variety of different tasks, including face attribute editing~\cite{choi2018stargan,he2019attgan,liu2019stgan,wu2019relgan,chusscgan2020,he2020pa,gao2021high,chen2020coogan,liu2021smooth}. Specifically, StarGAN~\cite{choi2018stargan} is the early work for learning multiple domain face translation, it takes multiple attributes as input, and it transfers one face image from one domain to other domains. During the training, StarGAN exploits a reconstruction and cycle consistency loss to preserve the content of the input face. After that, many works have been improving StarGAN, such as AttGAN~\cite{he2019attgan}, STGAN~\cite{liu2019stgan}, SSCGAN~\cite{chusscgan2020} and HifaFace~\cite{gao2021high}. However, their operation typically involves lower-resolution data (e.g., 256), limiting their capacity to manipulate 3D factors such as camera poses. On the other hand, our approach operates on higher-resolution data (e.g., 1024), enabling for more detailed manipulations and incorporating 3D factors seamlessly.

\noindent \textbf{Interpreting Latent Space of StyleGAN for Face Editing.} An alternative line of works explores the disentanglement of StyleGAN's latent space for face editing. These approaches can be roughly classified into two types based on whether they use semantic labels: unsupervised methods and attribute-conditional methods.
The former learns to discover interpretable directions in latent space by leveraging techniques such as Principal Component Analysis (PCA)~\cite{harkonen2020ganspace}, closed-form factorization~\cite{shen2021closed}, learnable orthogonal matrices~\cite{he2021eigengan,voynov2020unsupervised}, and regularization losses~\cite{peebles2020hessian,wei2021orthogonal}. GANSpace~\cite{harkonen2020ganspace} shows that PCA in the latent space of StyleGAN can find important interpretable directions that can be used to control image generation. To compute the PCA of the style codes, GANSpace samples multiple random vectors (i.e., $z$ space) and computes the corresponding style codes (\emph{i.e.}, $\mathcal{W}$ space). To avoid the extensive data sampling of GANSpace, SeFa~\cite{shen2021closed} directly decomposes the model weights with a closed-form solution. Similarly, recent works~\cite{he2021eigengan,voynov2020unsupervised} propose to obtain a disentangled latent space by learning an orthogonal matrix for editing latent code. 


As far as we know, the attribute conditions can be of different types, including global-level (\emph{e.g.}, label vectors) and local-level (\emph{e.g.}, semantic segmentation maps) modalities. The first type~\cite{abdal2021styleflow,liu2021isf,liang2021ssflow} usually utilizes off-the-shelf attribute classifier networks to obtain the attribute vectors of training images and then uses these vectors as input. For example, StyleFlow~\cite{abdal2021styleflow} proposes to utilize conditional normalizing flow (CNF) to model the mapping from conditional labels and latent codes ($z$ space) to intermediate vectors ($\mathcal{W}$ space). StyleFlow trains the flow model (CNF) with triplets consisting of vectors sampled from $\mathcal{W}$ space, corresponding faces, and predicted face attributes. Though StyleFlow can produce facial pose transformation, it suffers from serious view inconsistency as it lacks an understanding of the underlying 3D world. The second type utilizes the coarse masks~\cite{zhu2022region} or predicts face semantics~\cite{collins2020editing} with k-means.

Additionally, some methods~\cite{kwon2021diagonal,xu2022transeditor,shi2021semanticstylegan,kim2021exploiting} also explore the semantic disentanglement of the model, but they redesign the StyleGAN; thus, they need to retrain the generator. For example, TransEditor~\cite{xu2022transeditor} presents a transformer-based module for dual space interactions where one latent code is used as the key and value and the other as the query. This dual-space interaction helps disentangle the style and the content representations. Some works focus on local facial controls by integrating face parsing into generation~\cite{shi2021semanticstylegan} or by adding spatial information for styles code with the conv-based module~\cite{kim2021exploiting}.

Overall, these StyleGAN-based models have demonstrated the ability to produce high-quality images and perform precise editing. However, their limitations become evident when attempting to alter facial pose while maintaining view consistency. This challenge arises from their lack of robust 3D modeling abilities. However, our method overcomes this hurdle by inherently incorporating 3D modeling techniques (GNeRF), allowing for dynamic changes in facial pose while ensuring view consistency.

\noindent \textbf{3DMM-Guided Face Generation and Editing.} 
Recently, some works~\cite{geng20193d,deng2020disentangled,tewari2020stylerig,shi2021lifting,lin20223d} demonstrate high-quality control over GAN generation via a 3DMM~\cite{paysan20093d}. 3DMM is the 3D Morphable Face Model parameterized by the face shape, expression, and texture. For example, Geng \emph{et al.}~\cite{geng20193d} utilizes 3DMM to guide fine-grained face manipulation for arbitrary expression transfer. First, they extract texture and shape coefficients by fitting 3DMM to each real face in the dataset. Then, they utilize the texture generator to create the target textures with the source texture and the target expression and utilize the shape predictor to produce the target shape with the source shape coefficients and the target expression as input. Finally, the global generator utilizes rendered faces and the target expression to produce the final faces. StyleRig~\cite{tewari2020stylerig} and DiscoFaceGAN~\cite{deng2020disentangled} use 3DMM to manipulate the latent space of StyleGAN. While StyleRig is based on pre-trained StyleGAN models and only tunes a DFR module that learns the mapping from latent code to the coefficients of 3DMM.
On the other hand, DiscoFaceGAN re-trains the entire model. It exploits multiple VAE to model the distribution of 3DMM coefficients and introduces self-supervised losses to disentangle different factors. Compared to these models, our model does not require a 3DMM prior and still achieves better multi-view consistency. Additionally, our models can achieve more diverse face editing, such as hair color and age.

Finally, Shi \emph{et al.}~\cite{shi2021lifting} presents a LiftedGAN model, which lifts the pre-trained StyleGAN2 in 3D. This model is free of 3DMM prior. However, this model cannot achieve attribute-conditional control.

\noindent \textbf{GAN Inversion for Real Face Editing.}
GAN inversion aims to find an optimal latent code corresponding to a given real image and has been widely used for real image editing tasks. The previous methods can be divided into two broad categories: optimization-based~\cite{abdal2019image2stylegan,abdal2020image2stylegan++,roich2021pivotal} and encoder-based~\cite{zhu2020domain,richardson2021encoding,tov2021designing,alaluf2021restyle}. For example, Roich \emph{et al.}~\cite{roich2021pivotal} presents a novel optimization-based method called Pivotal Tuning Inversion (PTI). In PTI, they first obtain the optimized latent code as the pivot by fixing the parameters of the generator, then fix this pivot and finetune the generator parameters to obtain better reconstruction while preserving the editing abilities of the latent code. After the inversion step, they utilize the popular latent-disentanglement method, such as InterfaceGAN or GANSpace, for face editing. In this paper, we use the PTI for GAN inversion. Concurrent with our work, some methods~\cite{sun2022fenerf,sun2022ide,chen2022sem2nerf,jiang2022nerffaceediting} employ 3D-aware GAN as a basic model instead of StyleGAN to achieve multi-view consistent face editing guided by the segmentation masks. They also apply GAN inversion to project real images into latent space for editing. Different from these methods with segmentation masks, we use attribute labels to guide face editing.

\section{Preliminaries}

Our method can work with most of GNeRF backbones, and we showcase it with the two most recent ones, StyleSDF~\cite{orel2021stylesdf}, EG3D~\cite{chan2021efficient}. Since they have similar architecture, we only describe StyleSDF~\cite{orel2021stylesdf}. 

\subsection{Generative Neural Radiance Fields (GNeRF)} \label{gnerf}

NeRF~\cite{mildenhall2020nerf} is a continuous neural mapping $\mathcal{M}$ that maps a 3D position $\pmb{x}$ and a 2D viewing direction $\pmb{v}$ to the rgb color $\pmb{c}$ and density $\sigma$:
\begin{equation}
\begin{aligned}
(\pmb{c}, \sigma) = \mathcal{M}(\gamma(\pmb{x}),\gamma(\pmb{v})),
\end{aligned}
\end{equation}
where $\gamma$ indicates the positional encoding mapping function.  

GNeRF, such as GRAF~\cite{Schwarz2020NEURIPS}, is a conditional variant of NeRF. Unlike NeRF, which requires multiple views of a single scene with estimated camera poses, GNeRF can be trained with unposed 2D images from different scenes. In Pi-GAN~\cite{chanmonteiro2020pi-GAN}, GNeRF is trained with adversarial learning and conditioned on a latent code $\pmb{z}$:
\begin{equation}
\begin{aligned}
(\pmb{c}, \sigma) = \mathcal{M}(\gamma(\pmb{x}),\gamma(\pmb{v}), \pmb{z}),
\end{aligned}
\end{equation}
where the latent code $z$ with following MLP layers aims to infer frequencies $\alpha$ and shifts $\beta$ of a SIREN layer~\cite{chanmonteiro2020pi-GAN}. 

\noindent\textbf{StyleSDF}. As shown in Fig.~\ref{fig:model}, StyleSDF also adopts the SIREN layers inside GNeRF. However, it utilizes Signed Distance Fields (SDF) to improve the GNeRF and add a 2D StyleGAN generator as a second-stage rendering. In the first stage, the GNeRF is trained separately. It produces a feature vector $\pmb{V}$, RGB color $\pmb{c}$ and SDF values $\pmb{d}$:
\begin{equation}
\begin{aligned}
(\pmb{V},\pmb{c},\pmb{d}) = \mathcal{M}(\gamma(\pmb{x}),\gamma(\pmb{v}), \pmb{w}),
\end{aligned}
\end{equation}
where the learned SDF values define the object surface and thus allow the extract of the mesh via Marching Cubes~\cite{lorensen1987marching}. Moreover, $\pmb{d}$ will be converted into the density $\sigma$ for volume rendering.

The RGB color $\pmb{c}$ is later rendered into the low-resolution face image using the classical volume rendering. The output image is then used as input for the discriminator.

To achieve high-resolution output, the 3D volume feature $\pmb{V}$, undergoes volume rendering to produce a lower-resolution feature map, represented as $\pmb{V_{r}}$. Subsequently, $\pmb{V_{r}}$ is processed through Style-based convolutional modules to generate a high-resolution image. This high-resolution image is subsequently as input of another discriminator.


\section{The Proposed Method}

First, we detail the proposed `Training-As-Init, Optimizing-for-Tuning' (TRIOT) method. Next, we introduce a new method for the reference-based geometry transfer task.

\subsection{Training-as-Init, Optimizing-for-Tuning with Attribute-Specific Consistency Losses} 

As mentioned above, our proposed TRIOT has two stages: the training stage and the optimization stage. Given a pre-trained StyleSDF (or EG3D), we sample training data which includes lots of triplets: latent vector $\pmb{w}$ ($\mathcal{W+}$ space), the corresponding generated sample $\pmb{I}$ along with its low-resolution version $\pmb{I}_L$ and attribute labels $\pmb{A}$ predicted by \emph{off-the-shelf} attribute classifiers. During the training phase, the pre-trained StyleSDF model is augmented with a continuous normalized flows module. This module is trained using both latent vectors and corresponding labels. Subsequently, it facilitates the manipulation of latent codes, thereby enabling 3D-aware attribute editing during the inference phase. More details will be introduced below.

\noindent \textbf{Training Stage with Continuous Normalized Flows.}
We followed StyleFlow to use continuous normalized Flows (CNF) to learn the conditional distribution of latent code $\pmb{w}$. Normalized Flows are a type of generative model to map the prior distribution $\pmb{z} \thicksim p_{\pmb{z}}$ to a more complex distribution $\pmb{x} \thicksim p_{\pmb{x}}$, where the basic blocks of a normalizing flow is an invertible transformation $\pmb{x} = f(\pmb{z})$. Let define $f = f_{1} \circ f_{2}, ..., \circ f_{n}$, we have $\pmb{x}_{n} = f_{n} \circ, ..., f_{2} \circ f_{1}(\pmb{x}_{0})$. Then, it can be shown that $f$ is also bijective and invertible. Though a chain of mapping $f_{i}$, $p(\pmb{x}_{n})$ can be represented as using discrete normalized flows:

\begin{equation}
\begin{aligned}
\ln p_{n}(\pmb{x}_{n}) =  \ln p_{0}(\pmb{x}_{0})  - \sum\limits_{i = 1}\limits^{n} \ln \mid det \frac{df_{i}}{d\pmb{x}_{i - 1}}  \mid.
\end{aligned}
\end{equation}
where the function $f_{i}$ can be modeled by a neural network. 

StyleFlow uses the continuous normalized flows instead of the discrete normalized flows (DNF) to avoid the costly computation of Jacobian determinants in DNF while ensuring learning a reversible mapping. Specifically, the change in log density is:

\begin{equation}
\begin{aligned}
\ln p(\pmb{x}_{t_{i}}) =  \ln p(\pmb{x}_{t_{i-1}})  - \int^{t_{i}}_{\pmb{t_{i-1}}} Tr( \frac{dF}{d\pmb{x}_{t}}) dx ,
\end{aligned}
\end{equation}
where $t$ is the time variable in CNF. We use some CNF blocks to represent the function $F$, then train the CNF blocks using the latent code $\pmb{w}$ and the corresponding condition $\pmb{A}$. In the reference stage, we reversely inference the $\pmb{z}_{r}$ from $\pmb{w}_{r}$ using the original labels $\pmb{A}_{r}$ as the condition, then transfer $\pmb{w}_{r}$ to new latent code $\pmb{w}_{e}$ given target labels $\pmb{A}_{e}$ by the forward inference.

\begin{figure}[!t] \small
\centering
\includegraphics[width=1.0\linewidth]{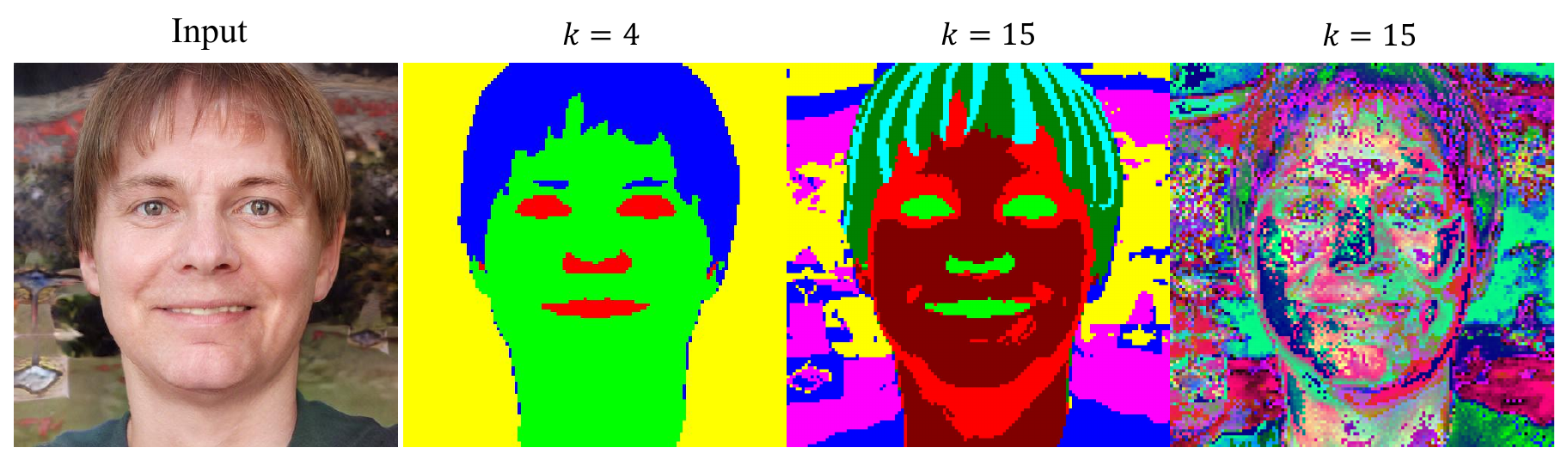}
\caption{Semantic decomposition involves applying clusters to the pretrained EG3D model. Here, $k$ represents the number of clusters utilized. The second and third columns display outcomes achieved through k-means clustering for rendering features $\pmb{S_{B}}$ from neural rendering. The final column exhibits rendering semantic maps generated via k-means clustering for 3D volumes $\pmb{S_{A}}$.}
\label{fig:clusters}
\vspace{-0.4cm}
\end{figure}

\begin{figure}[!t] \small
\centering
\includegraphics[width=1.0\linewidth]{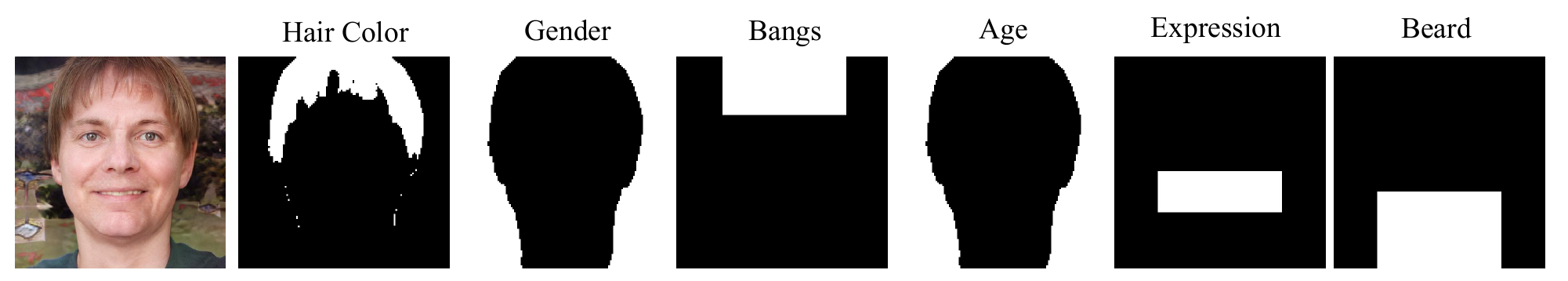}
\caption{Corresponding mask $M$ for all attributes.}
\label{fig:mask}
\vspace{-0.4cm}
\end{figure}

\noindent \textbf{Optimization Stage with Attribute-Specific Consistency Losses.}
Following training, latent codes are manipulated to generate images with altered attributes while maintaining view consistency. However, modifications to certain attributes can inadvertently affect unrelated attributes, particularly those that are local in nature. For instance, transferring a ``No-Smile'' face into a ``Smile'' can unintentionally alter the perceived identity of the individual. This phenomenon occurs due to the entanglement of latent codes for various attributes. To alleviate this problem, we propose an optimization stage to search for better latent codes corresponding to the better trade-off between the preservation of non-target regions and target-region editing. We propose an unsupervised method to extract attribute-specific masks to attain the target region, as shown in Fig.~\ref{fig:mask}. Motivated by previous feature visualization methods~\cite{collins2018deep,collins2020editing}, we explore applying spherical k-means clustering~\cite{hornik2012spherical} to the 3D Volume or rendering features of 3D GAN. Specifically, the tensor of 3D VOLUME is defined as $\pmb{V} \in R^{N\times C \times P \times H \times W}$, where $N$ is the number of images, $C$ is the dimension of channels, $P$ are sampling points of single ray, and $H$, $W$ are spatial dimensions. $\pmb{V}$ are rendered into features $\pmb{V_{r}} \in R^{N \times C \times H \times W}$. We firstly reshape $\pmb{V}$ and $\pmb{V_{r}}$ to $ \pmb{\hat V} \in R^{NHWP\times C}$ and $\pmb{\hat V_{r}} \in R^{NHW \times C}$, then apply the k-means for $\pmb{\hat V}$ and $\pmb{\hat V_{r}}$ to produce $k$ semantic maps.
\begin{equation}
\begin{aligned}
\pmb{S_{A}} &= kmeans(\pmb{\hat V}, k), \\
\pmb{S_{B}} &= kmeans(\pmb{\hat V_{r}}, k).
\end{aligned}
\end{equation}
Fig. \ref{fig:clusters} visualizes $\pmb{S_{A}}$ and $\pmb{S_{B}}$. We find that $\pmb{S_{A}}$ (4-th column) lacks specific meaningful semantics comparing to $\pmb{S_{B}}$ (3rd column). We choose $k=4$ (2nd column) for its more accurate semantic decomposition. Specifically, we utilize masks derived from $\pmb{S_{B}}$ for attributes like ``Hair Color'', ``Gender'' and ``Age''. For other localized attributes, we rely on pre-defined fixed rectangular masks for each facial image, given the well-aligned nature of the training faces. Fig.~\ref{fig:mask} shows all masks. With these masks, we define the non-target region as $\pmb{M}$ and target region as $1-\pmb{M}$. As shown in the right of Fig.~\ref{fig:model}, we obtain: $\pmb{I}_{r}$, $\pmb{I}^{n}_{r}$, $\pmb{I}_{e}$, $\pmb{I}_{e}^{n}$, and an initial latent code $\pmb{w}_{e}$. Given the corresponding optimization objection, we aim to find an optimal latent code $\pmb{\hat w}_{e}$, corresponding image $\pmb{\hat I}_{e}$ and normal map $\pmb{\hat I}_{e}^{n}$.

Specifically, our objective functions are attribute-specific, comprising mask-based texture consistency loss and mask-based geometry consistency loss. Illustrated on the right side of Fig.~\ref{fig:model}, our optimization objective for geometry ($\mathcal{L}_{mtloss}$) and texture ($\mathcal{L}_{mgloss}$) consistency is defined as follows:
\begin{eqnarray} \label{eq:consis}
        \mathcal{L}_{mtloss}&=&  \Vert \pmb{M} \odot (\pmb{\hat I}_{e} - \pmb{I}_{r}) \Vert_{1} \nonumber \\ &+& \Vert (1 - \pmb{M}) \odot ( \pmb{\hat I}_{e} - \pmb{I}_{e}) \Vert_{1}. \nonumber  \\
        \mathcal{L}_{mgloss}&=&\Vert \pmb{M} \odot (\pmb{\hat I}_{e}^{n} - \pmb{I}_{r}^{n}) \Vert_{1} \nonumber \\ &+& \Vert (1 - \pmb{M}) \odot ( \pmb{\hat I}_{e}^{n} - \pmb{I}_{e}^{n}) \Vert_{1}. \nonumber \\
\pmb{\hat w}_{e} &=& \argmin_{\pmb{\hat w}_{e}} \mathcal{L}_{mtloss} + \mathcal{L}_{mgloss}.
\end{eqnarray}
This texture loss ($\mathcal{L}_{mtloss}$) reduces differences between the reconstructed texture $\pmb{I}_{r}$ and the optimized texture $\pmb{\hat I}_{e}$ in the non-target region while keeping $\pmb{ \hat I}_{e}$ the same as $\pmb{I}_{e}$ in the target region for attribute-editing. The geometry consistency loss ($\mathcal{L}_{mgloss}$) has the same objective function as the texture loss. 


By iteratively optimizing the latent code $\pmb{\hat w}_{e}$, we can achieve enhanced editing results while effectively preserving non-target regions. This process typically requires fewer than 1,000 iterations, approximately taking two minutes.

\begin{figure}[!t]\small
\centering
\includegraphics[width=1.0\linewidth]{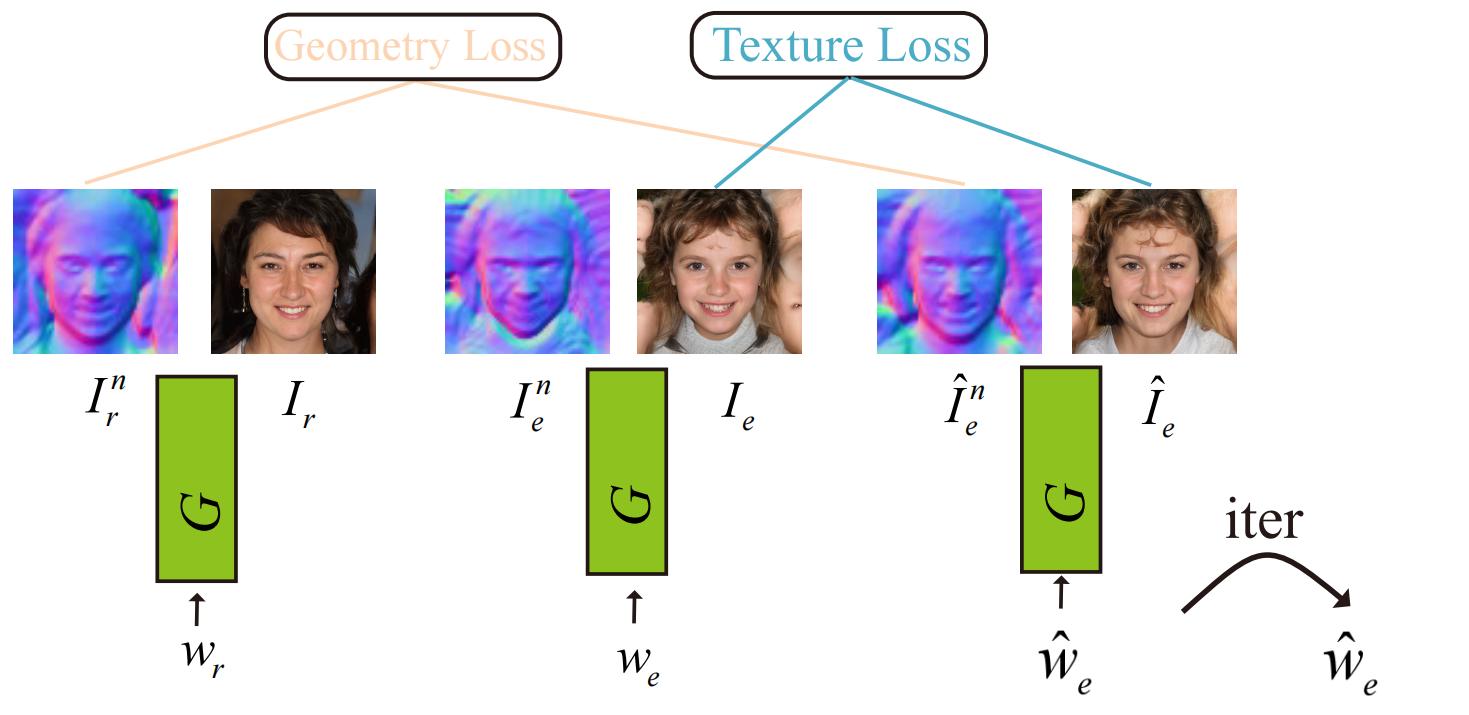}
\caption{Reference-based geometry transfer pipeline. It minimizes the difference between normals (geometry loss) and the differences between texture images (texture loss) in perceptual space to search for a better $\pmb{\hat w_{e}}$.}
\label{fig:model2}
\vspace{-0.4cm}
\end{figure}

\subsection{Reference-Based Geometry Transfer}  \label{geo}
Our model can achieve reference-based geometry manipulation by transferring the geometry from a reference face into a target face while preserving the target face's appearance. Like TRIOT, the optimization objection includes two parts: the geometry consistency loss and the texture consistency loss. Given a reference image and corresponding normal map, we minimize the differences between reference 
 and target normals (geometry loss) and the differences between the input image and target image (texture loss) in perceptual space. Specifically, we use $L1$ loss to compute the normal differences and $LPIPS$ metric to compute the texture differences in perceptual space. In addition, we incorporate an ID loss component aimed at preserving the unique identity features present in the target face. 

As shown in the Fig.~\ref{fig:model2}, given the original latent code $\pmb{w_{e}}$ and reference code $\pmb{w_{r}}$, the corresponding face image and normal pairs are ($\pmb{I}_{e}$,$\pmb{I}^{n}_{e}$)  ($\pmb{I}_{r}$, $\pmb{I}^{n}_{r}$), respectively. We use $\pmb{\hat w_{e}}$ as the target code of the optimization pipeline with corresponding face image and normal pairs ($\pmb{\hat I}_{e}$, $\pmb{\hat I}^{n}_{e}$). The optimization objective functions are defined as:

\begin{equation} \label{eq:ge}
\begin{aligned}
 \mathcal{L}_{tloss} &=  LPIPS(\pmb{I_{e}}, \pmb{\hat I_{e}}) + ID(\pmb{I_{e}}, \pmb{\hat I_{e}}),  \\
  \mathcal{L}_{gloss} &= \Vert (\pmb{I}^{n}_{r}  - \pmb{\hat I}^{n}_{e}) \Vert_{1}, \\
\pmb{\hat w_{e}} &=\argmin_{\pmb{\hat w_{e}}} \lambda_{1} \mathcal{L}_{tloss} +  \lambda_{2} \mathcal{L}_{gloss},
\end{aligned}
\end{equation}
where we employ a pre-trained face recognition network, ArcFace~\cite{deng2019arcface}, to calculate the ID loss for both images. This optimization phase efficiently discovers a latent code for precise reference-based geometry editing within just a few hundred steps.

\subsection{GAN Inversion}

We utilize the state-of-the-art GAN Inversion method (PTI) to project real images into the latent space of our 3D-GAN. Then, we perform real image editing using our proposed TRIOT methods. In detail, we follow two steps of PTI: 1) optimizing the latent code to obtain the corresponding projected latent of the real image while fixing the generator parameters (including CNF Blocks); 2) fixing the optimized latent code and the parameters of CNF parameters while tuning the remaining parameters of the generator. Then, given the generator's tuning parameters and projected latent code $\pmb{w}$ as the initial point, we used the proposed TRIOT method to search for better latent code following the texture and geometry consistency loss. Afterward, we can achieve precise image editing for given real images.

\section{Experiments}

We propose a method named TT-GNeRF (\textbf{T}raining and \textbf{T}uning \textbf{G}enerative \textbf{N}eural \textbf{R}adiance \textbf{F}ield). Within this framework, TT-GNeRF (S) and TT-GNeRF (E) denote the implementations of our approach using two different backbone networks: StyleSDF~\cite{orel2021stylesdf} and EG3D~\cite{chan2021efficient}, respectively. Additionally, we apply our method for StyleNeRF~\cite{gu2021stylenerf} and show results in the supplementary materials. We refer to the demo video~\footnote{https://ttgnerf.github.io/TT-GNeRF/} for more results on multiple-view attribute editing, reference-based geometry editing, and GAN inversion for real image editing.



\begin{figure*}[!t]\small
\centering
\includegraphics[width=0.98\linewidth]{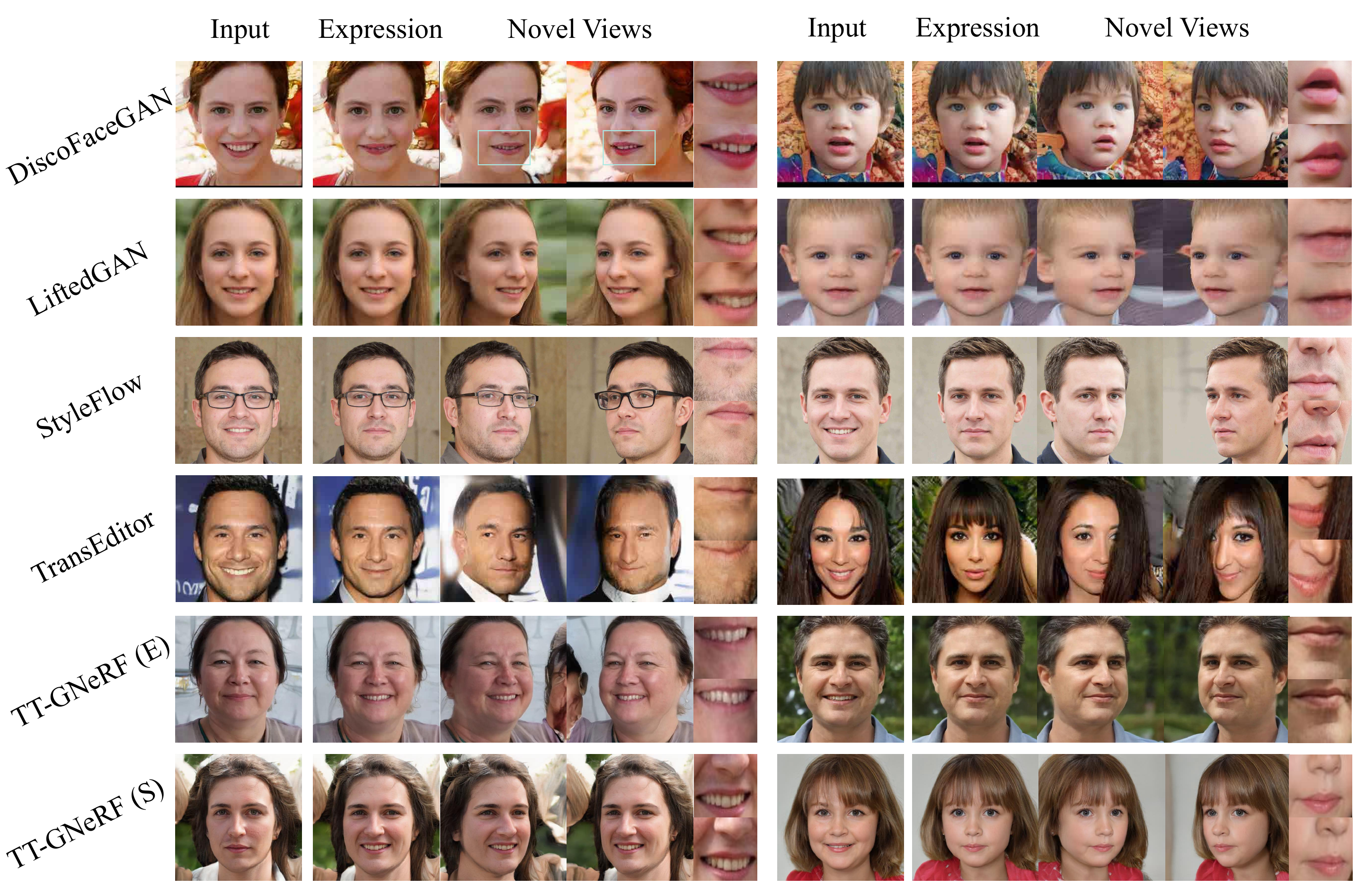}
\vspace{-0.2cm}
    \caption{Qualitative comparisons between our method and the baselines, \emph{i.e.}, DiscoFaceGAN~\cite{deng2020disentangled}, LiftedGAN~\cite{shi2021lifting}, StyeFlow~\cite{abdal2021styleflow}, and TransEditor~\cite{xu2022transeditor} on ``Expression'' attribute and the corresponding multiple-view renderings.}
\label{fig:exp2}
\vspace{-0.4cm}
\end{figure*}

\begin{figure*}[!t]\small
\centering
\includegraphics[width=0.98\linewidth]{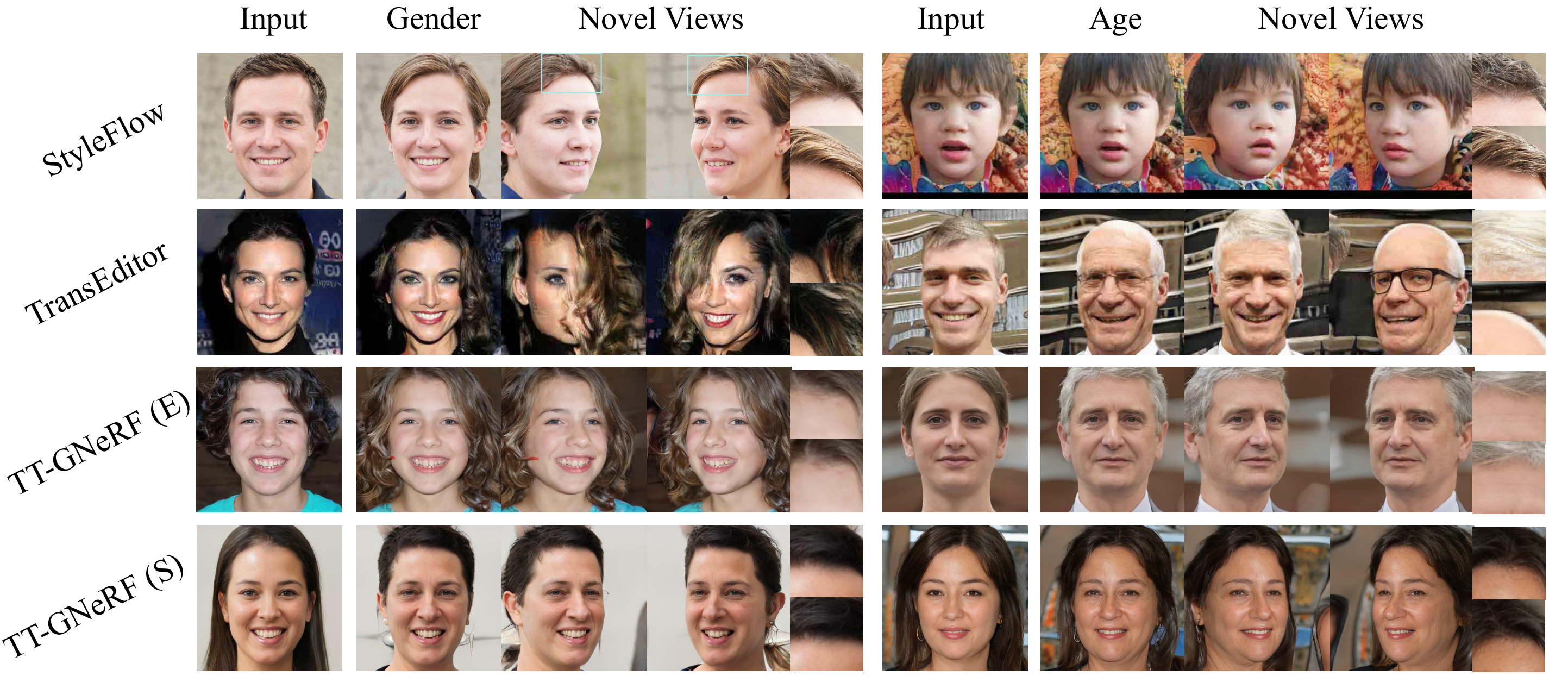}
    \caption{Qualitative comparisons between our method and the baselines, \emph{i.e.}, StyeFlow~\cite{abdal2021styleflow}, and TransEditor~\cite{xu2022transeditor} on ``Gender'' and ``Age'' attributes and the corresponding multiple-view renderings. Note that DiscoFaceGAN and LiftedGAN cannot deal with these two attributes.}
\label{fig:exp3}
\vspace{-0.4cm}
\end{figure*}

\begin{figure*}[!t]\small
\centering
\includegraphics[width=0.98\linewidth]{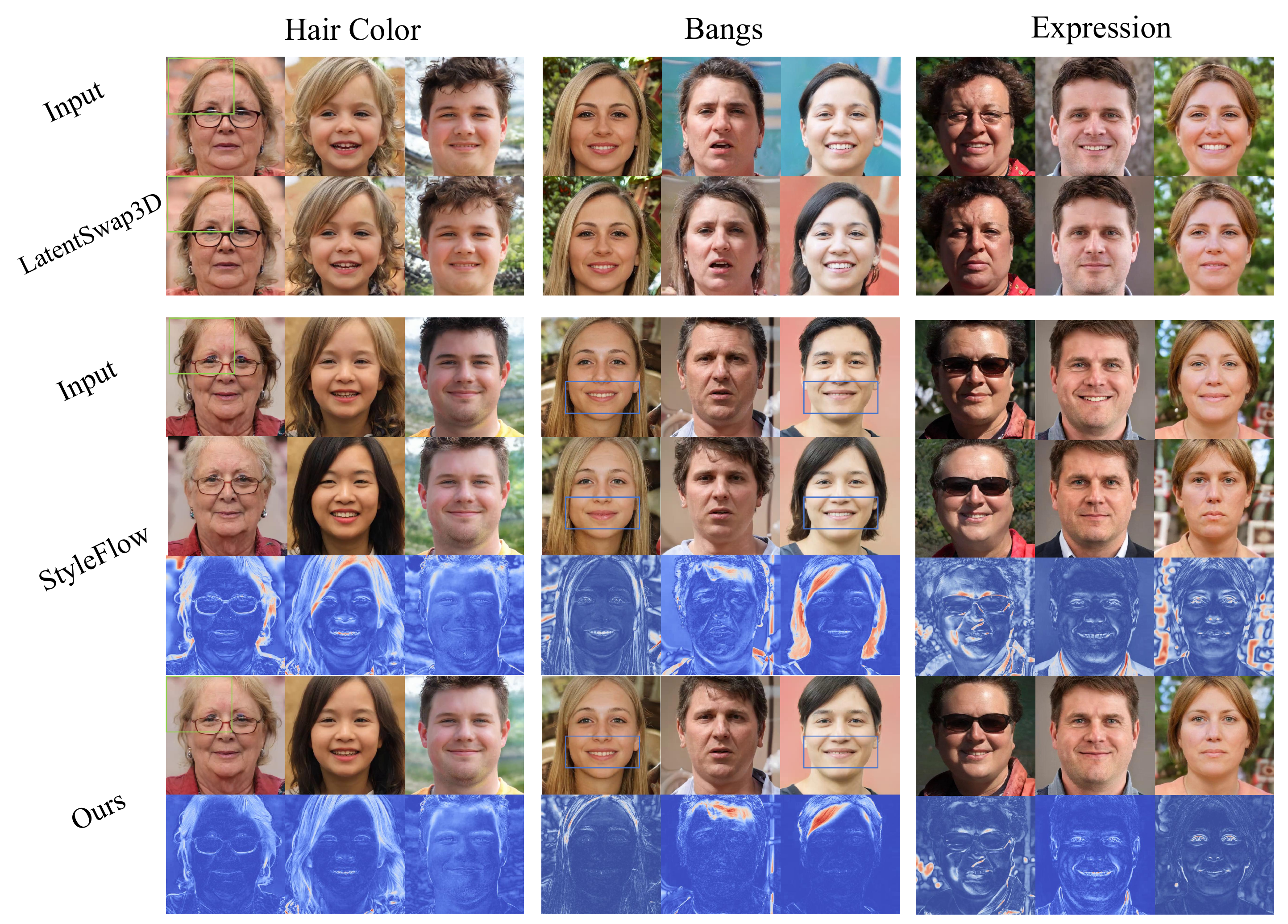}
    \caption{Qualitative comparisons with StyleFlow~\cite{abdal2021styleflow} and LatentSwap3D~\cite{simsar2023latentswap3d} based on EG3D.}
\label{fig:exp_styleflow}
\vspace{-0.4cm}
\end{figure*}

\begin{figure}[!t]\small
\centering
\includegraphics[width=0.98\linewidth]{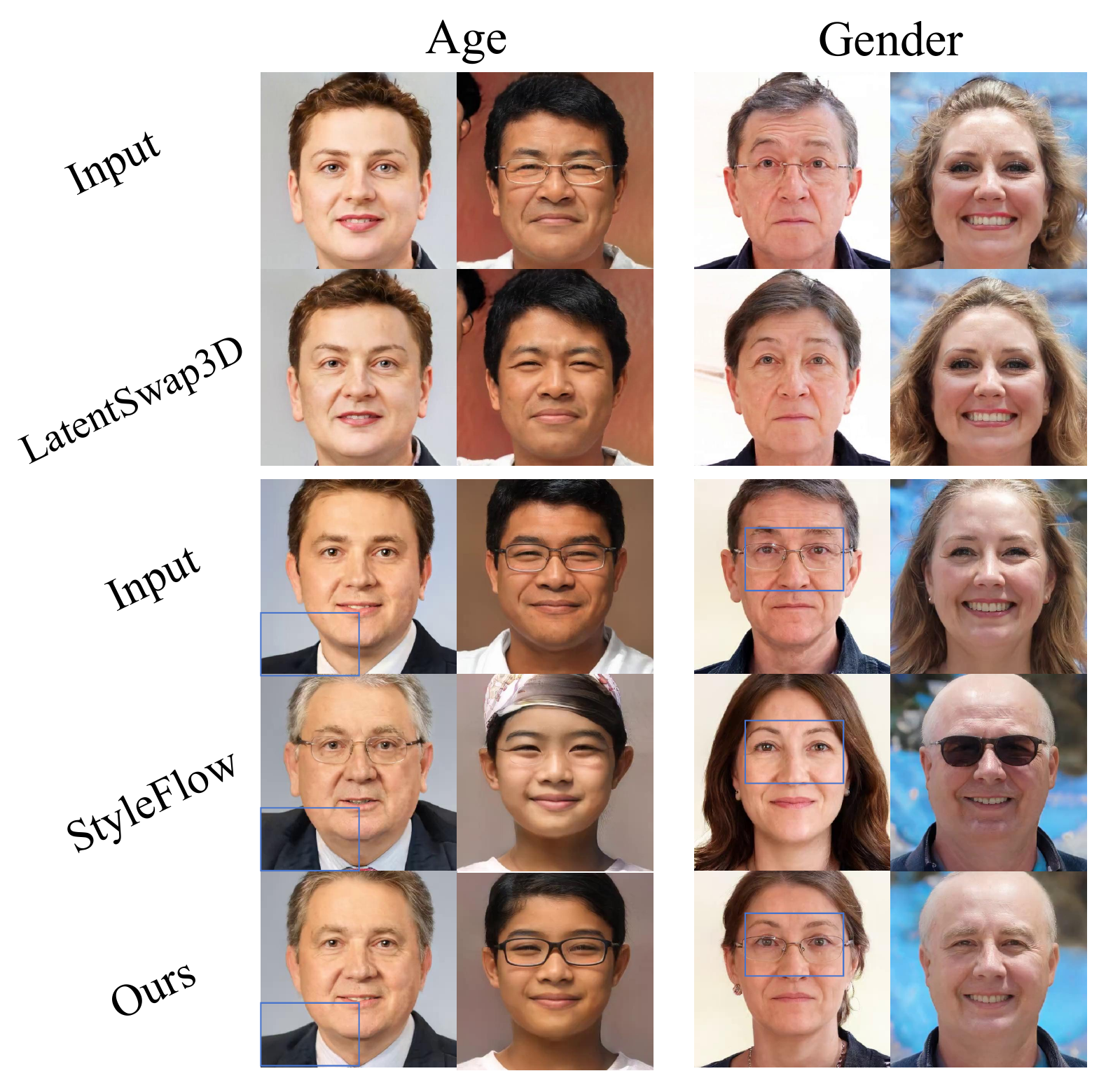}
    \caption{Qualitative comparisons with StyleFlow~\cite{abdal2021styleflow} and LatentSwap3D~\cite{simsar2023latentswap3d} based on EG3D.}
\label{fig:exp_styleflowv2}
\vspace{-0.4cm}
\end{figure}

\begin{table*}[t]
\caption{Quantitative results on the attributes editing results using four metrics: FID, Classification Accuracy (CA), Average Matching Point (aMP), and Face Recognition Similarity (FRS).}
\label{tab:q1}
\centering
	\resizebox{1\linewidth}{!}{
\begin{tabular}{lllllllllllll}
\toprule
\multirow{2}{*}{Method} & \multicolumn{4}{c}{Expression} &  \multicolumn{4}{c}{Gender} & \multicolumn{4}{c}{Age} \\
\cmidrule(r){2-5}
\cmidrule(r){6-9}
\cmidrule(r){10-13}
 & FID $\downarrow$ & CA $\uparrow$  & aMP $\uparrow$  & FRS $\uparrow$ & FID $\downarrow$ & CA $\uparrow$  & aMP $\uparrow$ & FRS $\uparrow$  &  FID $\downarrow$  & CA $\uparrow$ & aMP $\uparrow$ & FRS $\uparrow$  \\
\midrule
DiscoFaceGAN~\cite{deng2020disentangled} & 77.84 & 55.0 & 1347.4 & 0.587  & - & - & - & - & - & - & - & - \\
LiftedGAN~\cite{shi2021lifting} & 95.25 & - & 1484.0 & 0.464 & - & - & - & - & - & - & - & -  \\
StyleFlow~\cite{abdal2021styleflow} & 78.98 &  \bfseries 99.2 & 1089.7 & 0.586  & 82.80 & 79.9 & 1088.4 & 0.588 & 95.61 & 89.9 & 1090.8 & 0.586 \\
TransEditor~\cite{xu2022transeditor} & \bfseries 55.97 & 90.5 & 1075.6 & 0.564  & 76.73 & 96.3 & 964.30 & 0.575& 78.32 &  \bfseries 99.5 & 959.82 & 0.490    \\
\hline
TT-GNeRF (E) & 59.32 & 94.5 & 1529.5 & \bfseries 0.867 & 57.43  &  97.1 & 1532.7 & \bfseries 0.844 & 54.38 &  96.2 & 1643.5 & \bfseries 0.867  \\
TT-GNeRF (S) & 58.55 & 95.8 & \bfseries 1870.4 & 0.829 & \bfseries 54.78  & \bfseries 98.2 & \bfseries 1823.2 &  0.821 & \bfseries 55.28 & 90.5 & \bfseries 1974.6 & 0.834  \\
\bottomrule
\end{tabular}}
\vspace{-0.4cm}
\end{table*}

\subsection{Setting}

\begin{figure*}[!t]\small
\centering
\includegraphics[width=0.98\linewidth]{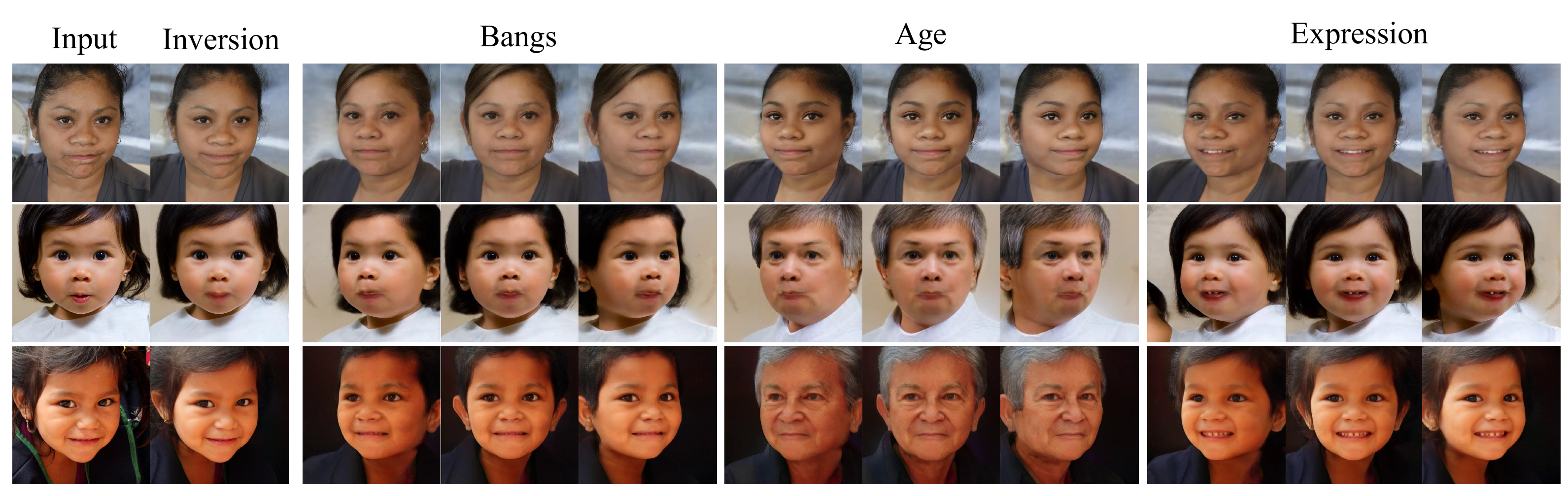}
    \caption{GAN Inversion for real image editing and corresponding multiple-view generation. The results are obtained using TT-GNeRF (E) with the target attributes of ``Bangs', ``Age'' and ``Expression''. For additional results, please refer to our demo video.}
\label{fig:exp_ap_4}
\vspace{-0.4cm}
\end{figure*}

\noindent \textbf{Datasets.} Given that the two backbones (\emph{i.e.}, StyleSDF and EG3D) are trained on the FFHQ dataset~\cite{karras2019style}, we train our model using sampled images and their corresponding latent codes. We create a dataset comprising 40,000 images at a resolution of $1024 \times 1024$ for training StyleSDF and an additional 40,000 images at a resolution of $512 \times 512$ for EG3D. We employ \emph{off-the-shelf} attribute-classifiers~\cite{karras2020analyzing} to obtain six attribute labels, including Hair Color, Gender, Bangs, Age, Expression (Smiling), and Beard. We use all generated triplets to train our models. The real images used for GAN inversion in Fig.~\ref{fig:exp_ap_4} are gathered from the Internet and subsequently aligned.

\noindent\textbf{Implementation Details.} During the training stage, CNF Blocks are trained using the Adam optimizer~\cite{kingma2014adam} with $\beta_{1}=0.9$, $\beta_{2}=0.999$, and a learning rate of $1e-4$ for 10,000 steps. In the optimization stage, we also utilize the Adam optimizer with $\beta_{1}=0.9$, $\beta_{2}=0.99$, and a learning rate of $5e-4$. Here, the number of steps is set within the range of (100 to 200) for all attributes. For the reference-based geometry-transfer task, we employ the Adam optimizer with $\beta_{1}=0.9$, $\beta_{2}=0.999$, and a learning rate of $5e-3$. The number of steps is fixed at 100. Both $\lambda_{1}$ and $\lambda_{2}$ in Eq.~\ref{eq:ge} are set to 1 and 10, respectively. 

\noindent\textbf{Compared Baselines.} Since our method is an attribute-conditioned generative model, the most similar supervised method is StyleFlow~\cite{abdal2021styleflow} and LatentSwap3D~\cite{simsar2023latentswap3d}. StyleFlow is a 2D method for face attribute editing tasks by manipulating the latent space of a 2D generator. To ensure a fair comparison, we extend the original Styleflow to work with 3D-aware generators, i.e., EG3D and StyleSDF. LatentSwap3D is a 3D method that is applied to multiple backbones, such as EG3D. Thus, we compare our method to LatentSwap3D based on the backbone EG3D. Moreover, we adopt the state-of-the-art generative model, TransEditor~\cite{xu2022transeditor} as our baseline for comparing face semantic disentanglement with multiple-view generation results. We also compare with the 3DMM-guided model, DiscoFaceGAN~\cite{deng2020disentangled}. Note that this model can only edit some expression-related attributes, such as ``Smile''. We adopt a 3D-Aware LiftedGAN~\cite{shi2021lifting} to compare multiple-view generation. However, LiftedGAN cannot control individual attributes. LENeRF~\cite{hyung2023local} is a very similar model to our TT-GNeRF but does not release the code. DragGAN~\cite{pan2023draggan} also optimizes a latent space of GAN to achieve image editing. We do not compare our method with DragGAN as it focuses on the semantic control of 2D GAN. We compare our method with StyleFlow~\cite{abdal2021styleflow} and GOAE~\cite{yuan2023make} for real image editing. GOAE is a recent 3D-aware image inversion and editing method.

\noindent\textbf{Evaluation Metrics.} Five metrics are used for evaluation: FID (Fréchet Inception Distance) score~\cite{NIPS2017_7240} to evaluate the quality and diversity of edited results; Classification Accuracy (CA) to evaluate the correctness of edited attributes; average Matching points (aMP)~\cite{zhang20213d} and Face Recognition Similarity (FRS)~\cite{liu2021isf} to quantitatively evaluate the consistency of multiple-view generation results; and Local Preservation (LP) to evaluate the preservation of non-target regions in editing results.

To evaluate the quality and diversity of edited results, we calculate the FID score~\cite{NIPS2017_7240} by using samples from FFHQ as the real distribution and the original image and its edited results as the fake distribution. To calculate FID scores, we sample 5000 real and fake samples from all models for each attribute. A lower FID score indicates a smaller discrepancy between the image quality of the real and generated images. 

To evaluate the accuracy of the attribute transfer, we use the \emph{off-the-shelf} classifiers~\cite{karras2020analyzing} to classify edited samples and compute the accuracy by comparing the predicted and target labels. We refer to this metric as Classification Accuracy (CA). We calculate the CA using 1000 edited samples from all models for each attribute. Higher CA indicates better accuracy.


Evaluating view consistency without ground truth is a challenging task. We address this challenge by using the average Matched Points metric~\cite{zhang20213d}. This metric involves computing a point-wise matching between two images ($I_1$, $I_2$) generated from the same identity but with different viewpoints using Patch2Pix~\cite{zhou2021patch2pix} and counting the number of Matched Points $MP(I_1, I_2)$. We calculate the mean of $MP$ across all pairs of samples for 100 random identities with ten views each to obtain a final {\em average MP} (aMP) score. Additionally, we use Face Recognition Similarity (FRS)~\cite{liu2021isf} to evaluate identity preservation across different views. Specifically, we use ArcFace~\cite{deng2019arcface}, a widely used face recognition method, to estimate feature similarity between two facial images and compute the average score across 1000 samples with ten different views and 100 identities. Higher aMP and FRS scores indicate that synthesized images with different viewpoints have more similar identities to input faces.

We use 1000 input-edited paired samples to evaluate the Local Preservation score (LP) for local attributes, such as ``Expression''. For every paired sample, we use the defined masks of Fig.~\ref{fig:mask} to attain the non-target region. Then, we measure the differences between each paired sample using $L1$ distance and average them across all pairs.

We have observed limitations in LiftedGAN and DiscoFaceGAN when directly manipulating attributes like ``Gender'' and ``Age'' for editing tasks. Consequently, we have opted not to assign scores for these methods in such cases. Instead, we have evaluated LiftedGAN based on FID, aMP, and FRS scores using randomly generated samples rather than attribute-edited ones.

\begin{table}[t]
\caption{Quantitative evaluation of the editing and preservation trade-off with the baselines, StyleFlow and LatentSwap3D, based on the EG3D. We use two metrics, CA and LP, to evaluate the editing of three attributes: Expression, Gender, and Age. Note that we define the non-target region by the mask of Fig.~\ref{fig:mask} to compute LP for three attributes.}
\label{tab:att_corr}
\centering
	\resizebox{1\linewidth}{!}{
\begin{tabular}{lllllll}
\toprule
\multirow{2}{*}{Method} & \multicolumn{2}{c}{Expression} &  \multicolumn{2}{c}{Gender} & \multicolumn{2}{c}{Age} \\
\cmidrule(r){2-3}
\cmidrule(r){4-5}
\cmidrule(r){6-7}
  & CA $\uparrow$  & LP $\downarrow$ & CA $\uparrow$  & LP $\downarrow$ & CA $\uparrow$ & LP $\downarrow$ \\
\midrule
StyleFlow~\cite{abdal2021styleflow} &   97.1 & 29.62 & 100 & 17.99 & 95.3 & 19.60 \\
LatentSwap3D~\cite{simsar2023latentswap3d} & 93.2 &  8.070 & 74.5 & 6.243 & 85.2 &  6.002 \\
TT-GNeRF (E) & 94.5 &  17.49 & 97.1 &  9.476 & 96.2 &  9.258 \\
\bottomrule
\end{tabular}}
\vspace{-0.4cm}
\end{table}

\begin{figure}[!t]\small
\centering
\includegraphics[width=0.98\linewidth]{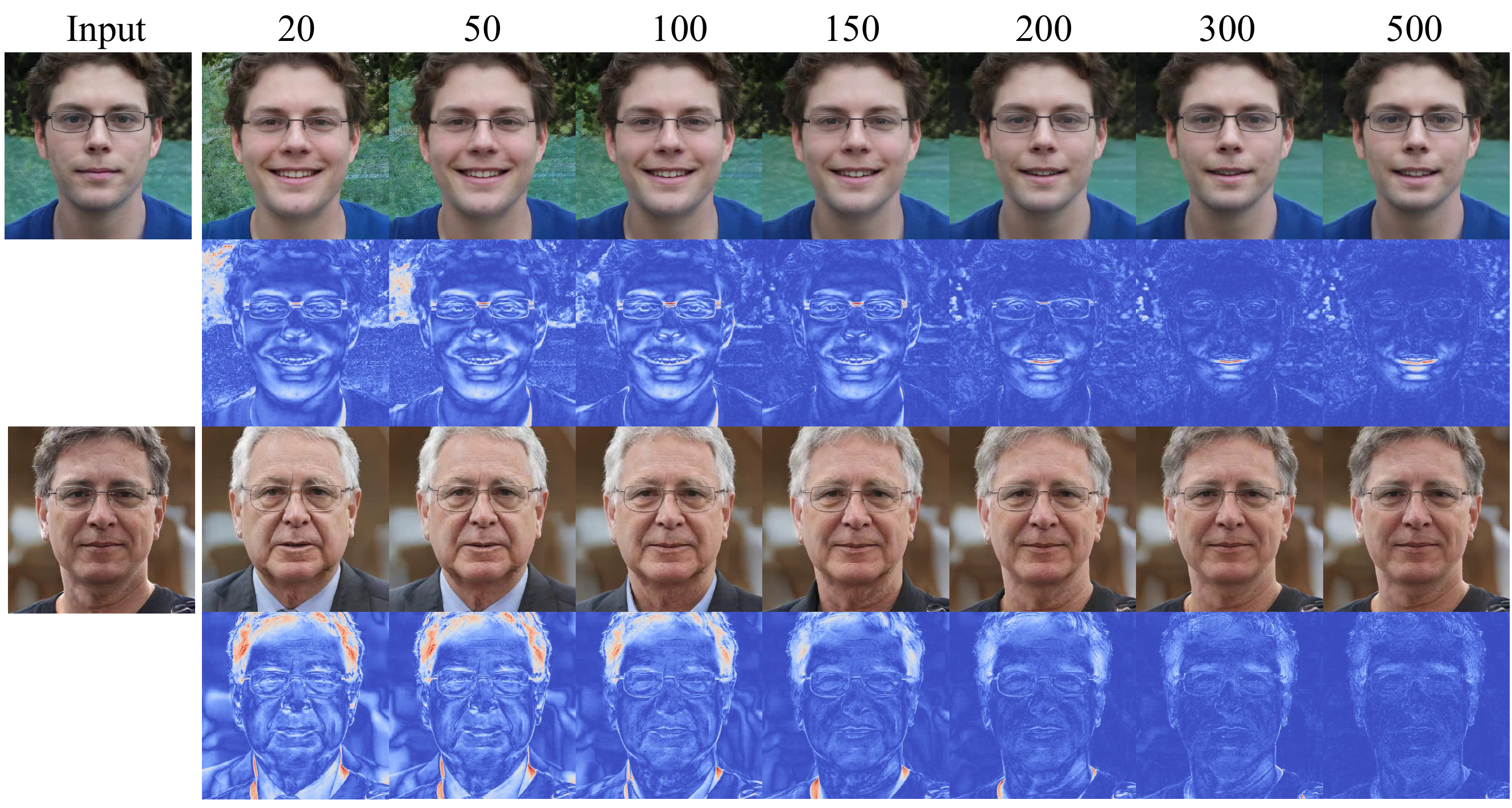}
    \caption{Ablations for the optimization steps. The attributes ``Expression'' and ``Age'' editing results are shown in the 2nd-8th columns.}
\label{fig:exp_ab2c}
\vspace{-0.4cm}
\end{figure}

\begin{figure}[t]\small
\centering
\includegraphics[width=0.98\linewidth]{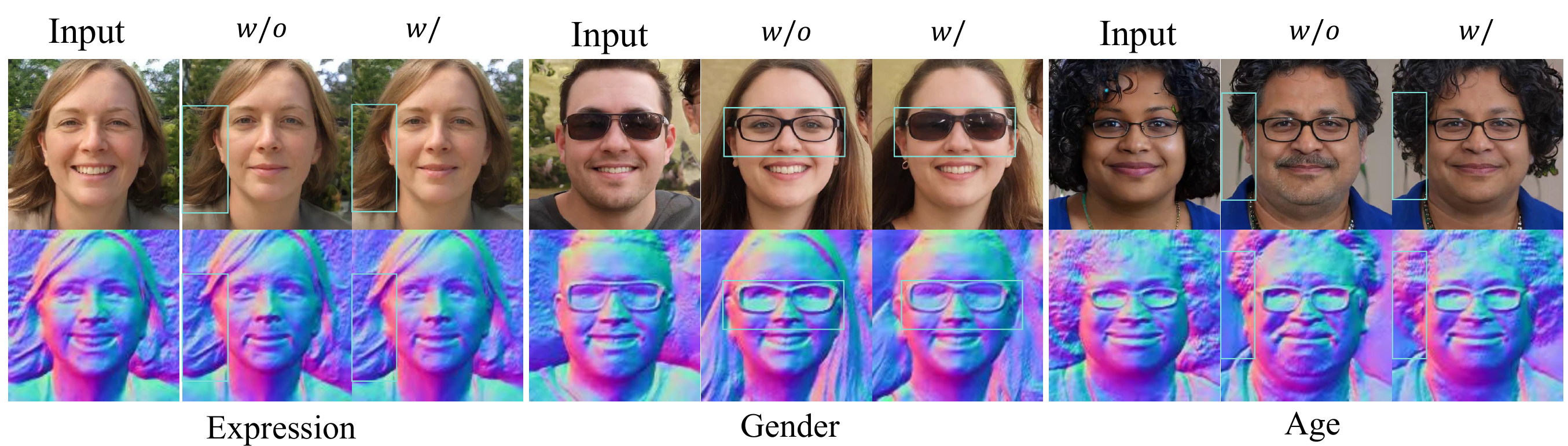}
    \caption{Ablations for the mask-based geometry consistency loss. For each attribute editing, the 2nd and second 3rd show the $w/o$ or $w/$ using the geometry consistency loss $\mathcal{L}_{mgloss}$ in the optimization stage, respectively.}
\label{fig:exp_ap_3}
\vspace{-0.4cm}
\end{figure}

\begin{figure*}[!t]\small
\centering
\includegraphics[width=0.98\linewidth]{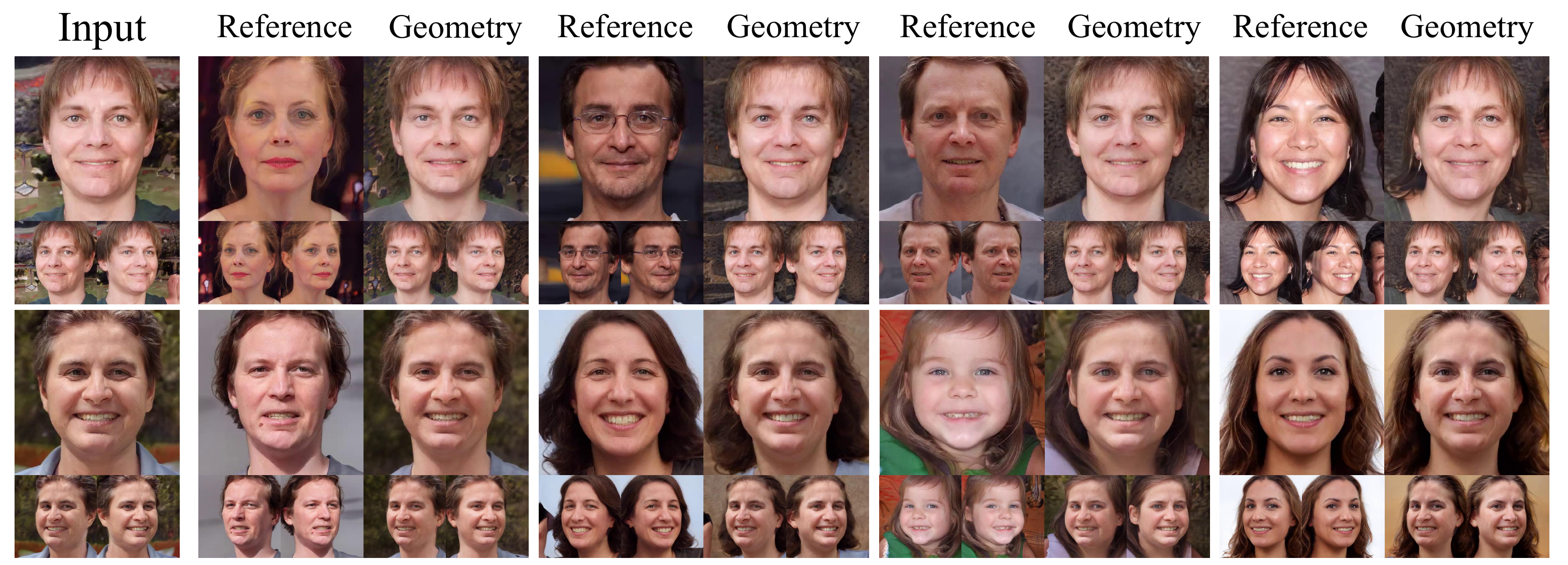}
    \caption{Visual results of reference-based geometry transfer from TT-GNeRF (E).}
\label{fig:exp_ap_2}
\vspace{-0.4cm}
\end{figure*}

\begin{figure}[h] 
\centering
\includegraphics[width=0.98\linewidth]{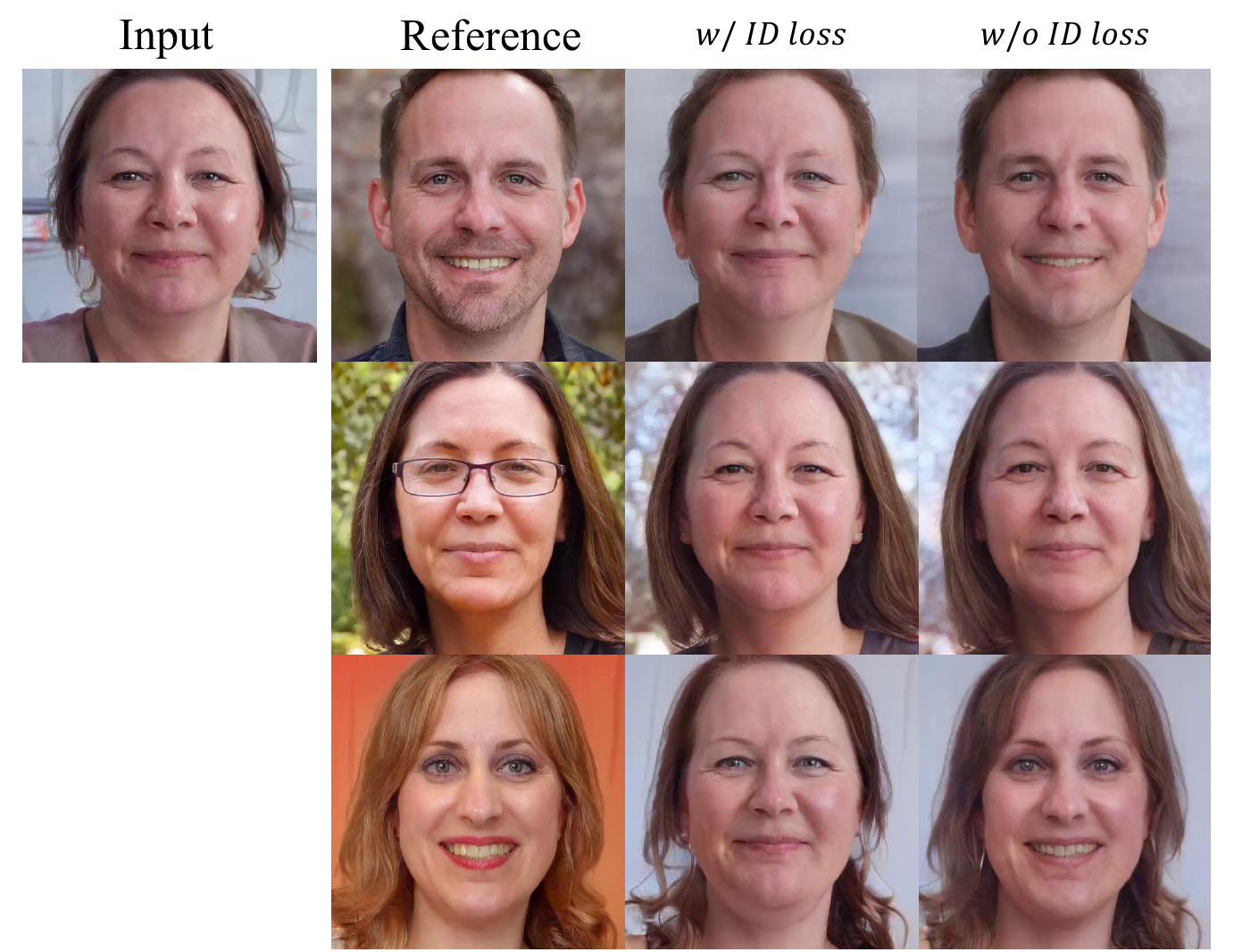}
    \caption{Ablation studies for ID loss of the reference-based geometry transfer method.}
\label{fig:exp_id}
\vspace{-0.4cm}
\end{figure}





\subsection{State-of-the-Art Comparison} \label{sec4.2}


We compare our method to 2D baselines on facial attribute editing with multiple-view generation in Fig.~\ref{fig:exp2} and Fig.~\ref{fig:exp3}. As aforementioned, we use ``Expression'', ``Gender'' and ``Age'' attributes as examples since most baselines can directly edit the three attributes. Fig.~\ref{fig:exp2} shows the results for ``Expression'' editing. We can observe that most models can accurately transfer between ``Smile'' and ``No-Smile''. However, TransEditor~\cite{xu2022transeditor} fails to preserve the non-target region, as shown in the 4-th row of Fig.~\ref{fig:exp2}. Moreover, our method outperforms all baselines in the 3D consistency for novel-view generation of edited results. For example, DiscoFaceGAN cannot maintain hair color when changing pose, and expression changes compared to the original view when zooming in on the mouth. StyleFlow and TransEditor struggle with large pose variations and suffer from severe view-inconsistency problems, resulting in significant identity changes such as beard growth on zoomed-in mouths. LiftedGAN demonstrates improved 3D consistency but exhibits limited quality and cannot perform facial attribute editing. Fig.~\ref{fig:exp3} illustrates the superiority of our method in 3D consistency compared to other approaches for ``Gender'' and ``Age'' editing. For detailed comparisons, please refer to the zoomed-in hair region.

\begin{table}
\caption{Quantitative ablation studies for the mask-based geometry consistency losses $\mathcal{L}_{mgloss}$. We report the $L1$ discrepancies of non-target regions between input and output samples, calculated across 50 textures (Tex) and normals (Norm).}
\label{tab:ab_geometryloss}
\centering
	\resizebox{1\linewidth}{!}{
\begin{tabular}{lllllll}
\toprule
\multirow{2}{*}{Method} & \multicolumn{2}{c}{Expression} &  \multicolumn{2}{c}{Gender} & \multicolumn{2}{c}{Age} \\
\cmidrule(r){2-3}
\cmidrule(r){4-5}
\cmidrule(r){6-7}
  & Norm $\downarrow$  & Tex $\downarrow$ & Norm $\downarrow$  & Tex $\downarrow$ & Norm $\downarrow$ & Tex $\downarrow$ \\
\midrule
$w/o$ $\mathcal{L}_{mgloss}$ & 6.10 &  5.42 & 4.20 &  16.23 & 3.78 &  10.25 \\
$w/$ $\mathcal{L}_{mgloss}$ & \bfseries 4.03 &  \bfseries 3.52 & \bfseries 3.38  & \bfseries 15.38  & \bfseries 3.29 & \bfseries 9.34 \\
\bottomrule
\end{tabular}}
\vspace{-0.4cm}
\end{table}

\begin{figure}[!t]\small
\centering
\includegraphics[width=0.98\linewidth]{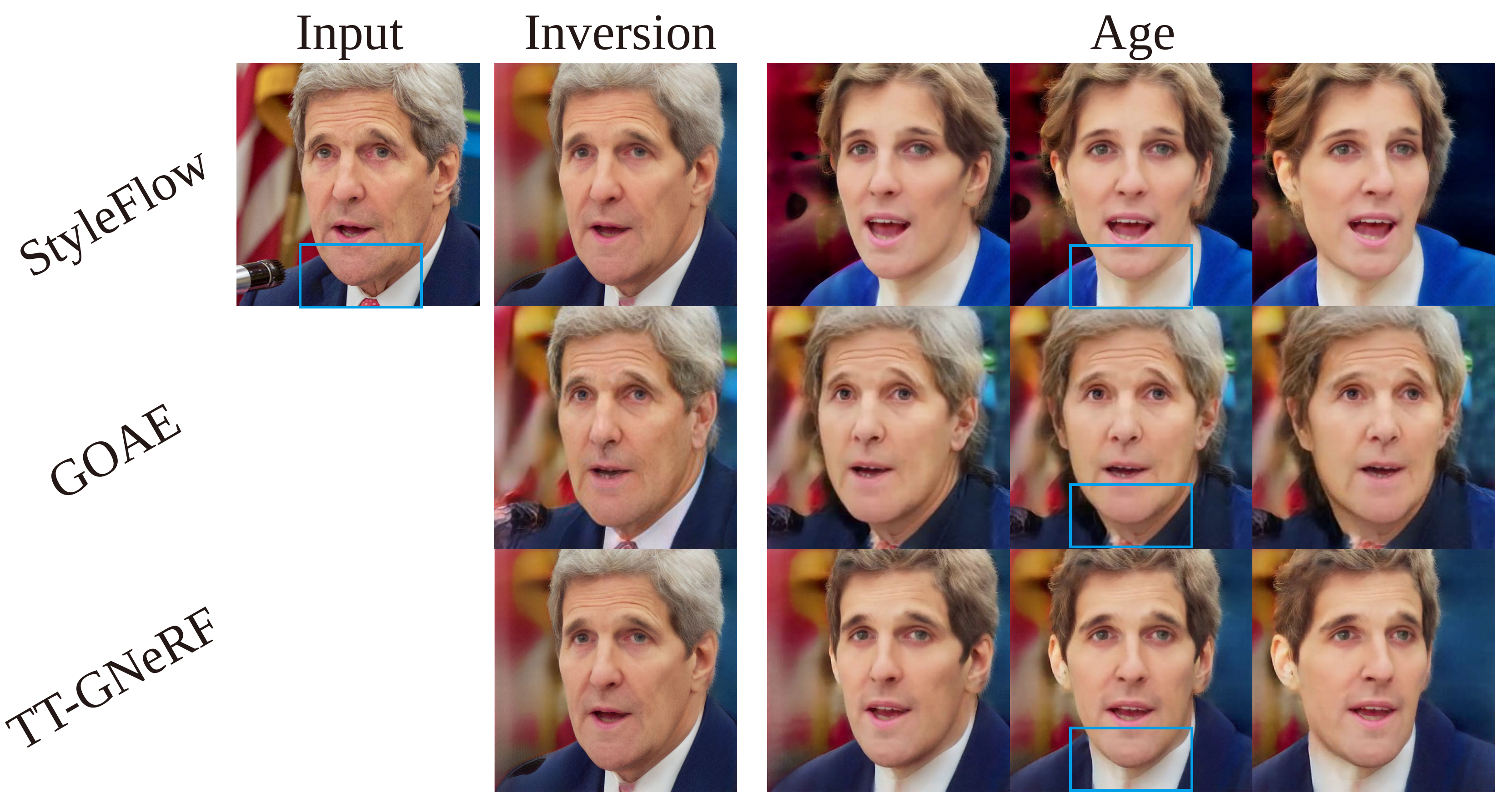}
    \caption{The comparison of the editing results between StyleFlow~\cite{abdal2021styleflow}, GOAE~\cite{yuan2023make} and TT-GNeRF (Ours).}
\label{fig:exp_ap_4v2}
\vspace{-0.4cm}
\end{figure}

\begin{figure}[!t]\small
\centering
\includegraphics[width=0.98\linewidth]{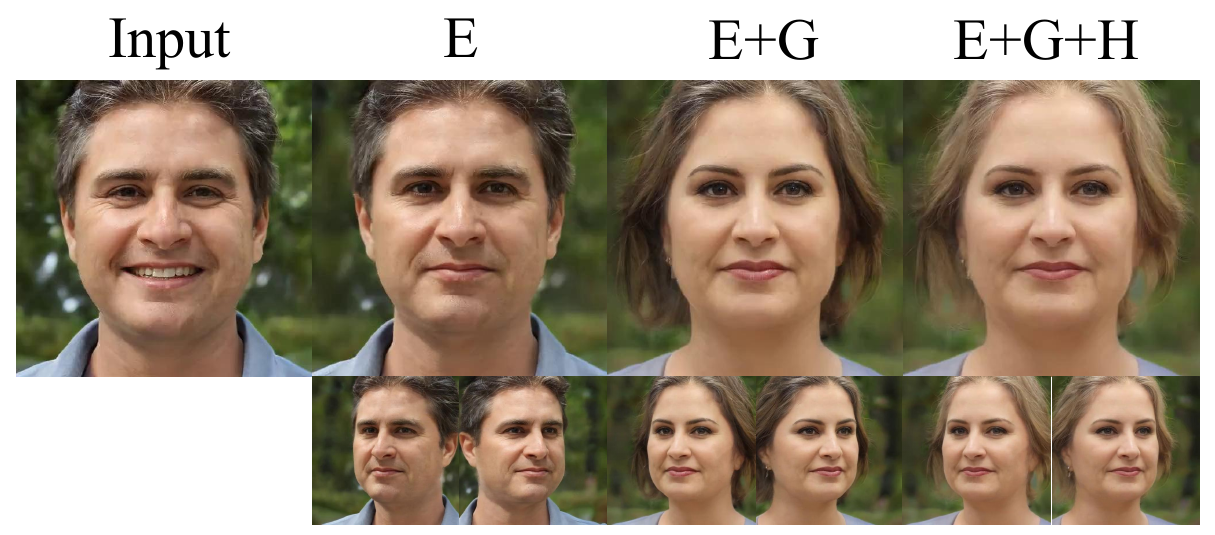}
    \caption{Multiple-attribute editing results from our TT-GNeRF (E) method. E: Expression, G: Gender, H: Hair Color.}
\label{fig:exp_ap_1}
\vspace{-0.4cm}
\end{figure}

Table~\ref{tab:q1} presents the quantitative evaluation results of ``Expression'', ``Gender'', and ``Age'' attributes editing. Our models achieve comparable performance to baselines for all three attributes regarding FID scores. Specifically, our TT-GNeRF (S) achieves the best FID scores for the ``Gender'' and ``Age'' attributes and a comparable score to TransEditor for the ``Expression'' attribute. Our models are also competitive with the baselines in terms of CA. For instance, TT-GNeRF (S) scores 98.2 for the ``Gender'' attribute, compared to 79.9 for StyleFlow and 96.3 for TransEditor. However, both models perform worse than the previous baselines for the ``Expression'' attribute. Table~\ref{tab:q1} shows that our models outperform previous methods on both aMP and FRS metrics for all three attributes. Specifically, our TT-GNeRF (S) achieves an aMP score of 1870.4 and an FRS score of 0.829 for the ``Expression'' attribute, outperforming LiftedGAN, which scores 1484.0 in aMP and 0.464 in FRS, as well as DiscoFaceGAN, which scores 1347.4 in aMP and 0.587 in FRS.

Fig.~\ref{fig:exp_styleflow} and Fig.~\ref{fig:exp_styleflowv2} compare the attribute editing results between TT-GNeRF and the baselines using the same 3D-aware generator (EG3D). Fig.~\ref{fig:exp_styleflow} shows the editing results for local attributes. We use the heatmap to visualize the input and editing results differences for TT-GNeRF and StyleFlow. For the attribute ``Bangs'', the 1st column shows that our TT-GNeRF can preserve other non-target regions, such as the mouth region, but StyleFlow significantly changes it (Blue Box). Similar results also can be found in the 3rd column for ``Bangs'' results. For other results, we can observe that our model can better preserve the identity or clothes than StyleFlow. Fig.~\ref{fig:exp_styleflowv2} shows the editing results for global attributes. We define the background and clothes as the non-target region for global attributes. For attribute ``Age'', the clothes of our editing results are more similar to input than StyleFlow (1st column (Blue Box)). For attribute ``Gender'', TT-GNeRF can preserve the clothes regions and other attributes (e.g., glasses (3rd column)), which validates the effectiveness of our proposed optimization stage.

For our comparative analysis between LatentSwap3D and our approach, we rigorously ensure a fair comparison by utilizing the same generator and latent code. Despite this, subtle divergences in identities between the two methods become apparent. Thus, to provide a comprehensive evaluation, we present side-by-side visual comparisons of inputs processed by our method and LatentSwap3D. Our findings reveal that LatentSwap3D suffers from very low editing intensity, particularly concerning attributes like ``Hair Color" (1st column (Green Box) of Fig.~\ref{fig:exp_styleflow}) and ``Gender" (4th column of Fig.~\ref{fig:exp_styleflowv2}). In contrast, our model showcases superior performance, striking a finer balance between effective editing and faithful preservation of the input attributes.

Our method emerges as the preferred choice, demonstrating enhanced editing precision and fidelity compared to LatentSwap3D and StyleFlow. This conclusion is supported by the quantitative results in Table~\ref{tab:att_corr}. In detail, TT-GNeRF has greatly improved LP scores for all attributes while having a CA score similar to StyleFlow. Moreover, our examination of LatentSwap3D reveals a notable trend: it exhibits a notably low CA score alongside a remarkably high LP score for both the "Gender" and "Age" attributes. This observation underscores a significant deficiency in editing intensity. 











\subsection{Ablation Study}

In the following, we present ablation studies on the effect of the geometry consistency loss, optimization steps in the optimization stage, and ID loss of the reference-based geometry editing.

\noindent\textbf{Geometry Consistency Loss.}  Our proposed TRIOT can further improve identity preservation after face attribute editing. In the optimization stage, the objective function (Eq.~\ref{eq:consis}) includes the mask-based texture consistency loss and the mask-based geometry consistency loss. We compare the full objective function with one ablation baseline, which removes the geometry losses. As shown in Fig.~\ref{fig:exp_ap_3}, we compare two methods by the attribute editing results. Our method excels in preserving non-target attributes for three attributes, particularly within the designated rectangular region (e.g., eyeglasses for attribute ``Gender''). Notably, when examining the normals (2nd row), it becomes evident that the ``without'' ($w/o$) method fails to circumvent geometric alterations, whereas the full method adeptly sidesteps this issue. Table~\ref{tab:ab_geometryloss} reports the $L1$ differences between input and editing results in the non-target region, providing a quantitative assessment of the effectiveness of the proposed geometry consistency loss.

\noindent\textbf{Optimization Steps.} We investigate how the optimization stage steps impact the balance between editing accuracy and identity (or background) preservation. As depicted in Fig.~\ref{fig:exp_ab2c}, we notice the smiled and aged results have very high transferring intensity in the few steps setting. However, as the optimization steps increase, the aged and smiled results progressively converge towards the input. Therefore, we choose a configuration of (100 and 200) steps, representing the optimal balance between precision in attribute editing and identity preservation.

\noindent\textbf{Identity Loss (ID Loss).} We investigate the impact of identity loss in reference-based geometry editing tasks. As illustrated in Fig.~\ref{fig:exp_id}, it is evident that employing identity loss ($w/$ ID loss) results in superior preservation of the input image's identity compared to the absence of identity loss ($w/o$ ID loss). For example, in the first row, $w/o$ ID loss fails to maintain the gender of the input image.

\subsection{Applications}

\noindent \textbf{Reference-based Geometry Transfer.}  
Disentangling geometry and textures has proven challenging in prior methods. We introduce an optimization approach for reference-based geometry transfer. As depicted in Fig.~\ref{fig:exp_ap_2}, our method yields promising results in geometry transfer, accompanied by corresponding multi-view outcomes. Our method effectively integrates textures from input images into the geometry of the reference image, resulting in highly realistic outputs.

\noindent \textbf{GAN Inversion for Real Image Editing.} 
Fig.~\ref{fig:exp_ap_4} shows real image inversion and editing results for ``Bangs'', ``Age'' and ``Expression'' attributes. The 2-th column shows we can produce almost perfect reconstruction for the real images. After that, our model can achieve accurate attribute editing. Fig.~\ref{fig:exp_ap_4v2} shows the inversion and editing results of StyleFlow, GOAE, and ours. We observe that the non-target region, such as the clothes, is well preserved for our method.
 
\emph{We refer to the demo video for more results about multiple-view attribute editing, reference-based geometry editing, and GAN inversion for real image editing.}

\noindent \textbf{Multiple-attribute Editing.} 
In addition to the previous single-attribute edits, our model can perform sequential editing of multiple attributes. Fig.~\ref{fig:exp_ap_1} shows high-quality edits for the sequence ``Expression + Gender + Hair color''. Multiple-view results for these edits are presented at the bottom of each row, demonstrating that our model maintains strong 3D-view consistency in these cases. 

\section{Conclusions}

In this work, we propose an attribute-conditional 3D-aware face generation and editing model, which shows the disentangling abilities of the generative neural radiance field with labels as inputs. Moreover, we integrate the training method with normalized flows and the TRIOT optimization method into the 3D-aware face editing models to achieve the best balance between attribute-editing precision and non-target attribute preservation. The qualitative and quantitative results demonstrate the superiority of our method. Additionally, we explore the disentanglement of the geometry and textures by achieving reference-based geometry transfer task with the elegant optimization method while preserving the appearance. However, there still exist some limitations. First, our model fails to edit the facial attribute in some cases. For example, the expression part of Table~\ref{tab:q1} shows that our CA score is worse than some methods. Second, our proposed TRIOT still costs some minutes for single attribute editing; thus, it is unacceptable for some real application scenarios. Finally, as shown in Fig.~\ref{fig:exp_ap_4}, our model can achieve the single image 3D model and perform attribute editing. However, compared to video-based head avatars~\cite{gafni2021dynamic,zheng2022avatar}, the identity is not well preserved between real images and projected images. Proposing better GAN inversion techniques adapted for 3D-Aware GAN can further alleviate this problem. 

\bibliographystyle{IEEEtran}
\bibliography{main}

\end{document}


\title{Training and Tuning Generative Neural Radiance Fields for Attribute-Conditional 3D-Aware Face Generation--Supplementary Material--}



\author{Jichao~Zhang,
        Aliaksandr~Siarohin,
        Yahui~Liu,
        Hao~Tang, 
        Nicu~Sebe,~\IEEEmembership{Senior~Member,~IEEE},
        and Wei Wang
\IEEEcompsocitemizethanks{\IEEEcompsocthanksitem Jichao Zhang is with the School of Computer Science, Ocean University of China, Shandong, China. E-mail: zhang163220@gmail.com. 
\IEEEcompsocthanksitem  Aliaksandr Siarohin is with the Snap Research, Santa Monica, CA,
US. E-mail: aliaksandr.siarohin@gmail.com.
\IEEEcompsocthanksitem  Yahui Liu is a Principal Engineer in Huawei, Shenzhen, China. E-mail: yahui.cvrs@gmail.com.
\IEEEcompsocthanksitem Hao Tang is with the School of Computer Science, Peking University, Beijing, China. E-mail: bjdxtanghao@gmail.com.
\IEEEcompsocthanksitem Nicu Sebe is with the Department of Information Engineering and Computer Science (DISI), University of Trento, Italy. E-mail: sebe@disi.unitn.it. 
\IEEEcompsocthanksitem Wei Wang is with the Institute of Information Science, Beijing Jiaotong University, Beijing, China. E-mail: wangwei1990@gmail.com.}
}

\markboth{Submitted to IEEE Transactions on Visualization and Computer Graphics}%
{Shell \MakeLowercase{\textit{et al.}}: A Sample Article Using IEEEtran.cls for IEEE Journals}


\maketitle

\begin{IEEEkeywords}
Neural Radiance Fields, Generative Model, Generative Adversarial Networks, Image Generation and Editing.
\end{IEEEkeywords}

\section{More Ablation Studies}

\noindent\textbf{Comparison with StyleFlow based on StyleSDF.} 
Fig.~\ref{fig:exp_stylesdf} illustrates the comparison between our comprehensive method and StyleFlow, utilizing the same 3D generator, StyleSDF. Our analysis reveals that our approach attains a superior balance between editing capabilities and preservation of the non-target region.

\begin{figure}[t]
\centering
\includegraphics[width=0.98\linewidth]{pdf/figure_stylesdf.pdf}
    \caption{Qualitative comparisons based on StyleSDF~\cite{orel2021stylesdf}.}
\label{fig:exp_stylesdf}
\vspace{-0.4cm}
\end{figure}

\noindent\textbf{Geometry Consistency losses.} 
We conducted ablation studies on the effectiveness of the geometry consistency loss in the optimization stage, as presented in the main paper's Fig. 11. Here, we extend our comparison to include ``Hair Color'' attributes as illustrated in Fig.~\ref{fig:exp_geo}.

\begin{figure}[t]
\centering
\includegraphics[width=0.98\linewidth]{pdf/figure_ab_normal.pdf}
    \caption{Qualitative ablation studies of the mask-based geometry consistency loss (Eq.7). Taking attribute ``Hair Color'' as an example.}
\label{fig:exp_geo}
\vspace{-0.4cm}
\end{figure}

\noindent\textbf{Normal Losses.} 
In our optimization process, we utilize the L1 loss as the objective function for normals. Here, we delve into comparing three different losses: L1, L2, and Angular. Fig.~\ref{fig:exp_angular} showcases the editing outcomes for two attributes alongside their corresponding normals, accompanied by visual heatmaps illustrating the disparities between each editing result and the input image. These findings demonstrate remarkably similar results across all methods employing these diverse loss functions. Consequently, we have opted to maintain the L1 loss as our optimization objective function.

\begin{figure}[t]
\centering
\includegraphics[width=0.98\linewidth]{pdf/figure_ab_angle_loss.pdf}
    \caption{Ablation studies of different normal losses: L1, L2, and Angular losses.}
\label{fig:exp_angular}
\vspace{-0.4cm}
\end{figure}

\section{A new backbone: StyleNeRF}

\noindent\textbf{Training Details for StyleNeRF.}
Given that the backbone StyleNeRF~\cite{gu2021stylenerf} is trained on the FFHQ dataset~\cite{karras2019style}, we train our model using sampled images and their corresponding latent codes. We employ a dataset comprising 10,000 images at a resolution of $256 \times 256$ for training StyleNeRF. We employ \emph{off-the-shelf} attribute-classifiers~\cite{karras2020analyzing} to obtain six attribute labels, including Hair Color, Gender, Bangs, Age, Expression (Smiling), and Beard. We use all generated triplets to train our models. 

\noindent\textbf{Results.}
Fig.~\ref{fig:exp_stylesdf}  shows the comparison with StyleFlow, and Fig.~\ref{fig:exp_stylesdfv2} shows diverse attribute editing results from our method.

\begin{figure}[t]
\centering
\includegraphics[width=0.98\linewidth]{pdf/figure_stylenerf.pdf}
    \caption{Qualitative comparisons between our method and StyleFlow based on StyleNeRF~\cite{gu2021stylenerf}.}
\label{fig:exp_stylesdf}
\vspace{-0.4cm}
\end{figure}

\section{Unsupervised Semantic Decomposition}
We have delved into an unsupervised technique to extract semantic masks from rendering features generated by a pre-trained 3D-aware GAN model. Our approach involves utilizing clustering methods on the rendering features of the 3D-aware GANs to derive semantic masks for attributes such as "Background," "Gender," "Age," and "Hair Color." The rendering feature dimension and $K$ for EG3D are 128 and $4$; The rendering feature dimension and selected $K$ for StyleSDF are 64 and $3$; The selected feature dimension and $K$ for StyleNeRF are 128 and $3$. Note that we select a feature layer with $128\times128$ dimension from the super-resolution module of StyleNeRF.

\noindent\textbf{Results.}
Fig.~\ref{fig:semantics} shows some extracted semantic masks based on two backbones, EG3D and StyleSDF. In our experiments, we set the numbers of clusters $k=4$ for EG3D and $k=3$  for StyleSDF.

\begin{figure}[t]
\centering
\includegraphics[width=0.98\linewidth]{pdf/figure_com_faceparsing.pdf}
    \caption{Comparison of Mask Utilization: Our Unsupervised K-Means Method versus Face Parsing Model~\cite{yu2018bisenet}.}
\label{fig:semantics}
\vspace{-0.4cm}
\end{figure}

\begin{table}[t]
\caption{Quantitative comparison of Mask Utilization: Our Unsupervised K-Means Method versus Face Parsing Model~\cite{yu2018bisenet}.}
\label{tab:masks}
\centering
	\resizebox{1\linewidth}{!}{
\begin{tabular}{llllll}
\toprule
\multirow{2}{*}{Method} & \multicolumn{2}{c}{Age} &  \multicolumn{2}{c}{Gender} \\
\cmidrule(r){2-3}
\cmidrule(r){4-5}
  & CA $\uparrow$  & LP $\downarrow$ & CA $\uparrow$  & LP $\downarrow$ \\
\midrule
Face Parsing Model~\cite{yu2018bisenet} & 97.5 &  8.002 & 98.0 & 9.104 \\
K-Means & 96.2 &  9.258 & 97.1 &  9.476 \\
\bottomrule
\end{tabular}}
\vspace{-0.4cm}
\end{table}

\begin{figure*}[t]
\centering
\includegraphics[width=0.98\linewidth]{pdf/figure_stylenerfv2.pdf}
    \caption{More attribute editing results from our method based on StyleNeRF~\cite{gu2021stylenerf}.}
\label{fig:exp_stylesdfv2}
\vspace{-0.4cm}
\end{figure*}

\begin{figure*}[t]
\centering
\includegraphics[width=0.98\linewidth]{pdf/figure_semantics.pdf}
    \caption{More Semantic decomposition results based on EG3D and StyleSDF.}
\label{fig:semantics}
\vspace{-0.4cm}
\end{figure*}

\subsection{Comparison to off-the-thelf face parsing tools.}

We introduce an unsupervised approach utilizing cluster methods (K-means) to extract semantic masks for attributes such as ``Gender'', ``Age'', and ``Hair Color''. Fig.~\ref{fig:semantics} presents a qualitative comparison with an off-the-shelf face parsing model for extracting attribute masks. Both methods yield similar attribute editing results, although the masks generated by the k-means method are less precise compared to the face parsing model. The quantitative results in Table~\ref{tab:masks} further confirm the similarity in performance between both methods.

\bibliographystyle{IEEEtran}
\bibliography{tvcg}
